\documentclass{article}

\usepackage{PRIMEarxiv}

\usepackage[utf8]{inputenc} % allow utf-8 input
\usepackage[T1]{fontenc}    % use 8-bit T1 fonts
\usepackage{hyperref}       % hyperlinks
\usepackage{url}            % simple URL typesetting
\usepackage{booktabs}       % professional-quality tables
\usepackage{amsfonts}       % blackboard math symbols
\usepackage{nicefrac}       % compact symbols for 1/2, etc.
\usepackage{microtype}      % microtypography
\usepackage{lipsum}
\usepackage{fancyhdr}       % header
\usepackage{graphicx}       % graphics
\graphicspath{{media/}}     % organize your images and other figures under media/ folder
\usepackage{cite}
\usepackage{amsmath,amssymb,amsfonts}
\usepackage{algorithmic}
\usepackage{graphicx}
\usepackage{textcomp}
\usepackage{multirow}
\usepackage{makecell}
\usepackage{subfigure}
\usepackage{colortbl}
\usepackage{xcolor}
\usepackage{booktabs}
\usepackage{arydshln}
\usepackage[export]{adjustbox}
\usepackage[lined,boxed,commentsnumbered,ruled]{algorithm2e}
\begin{document}
\title{Pan-cancer Histopathology WSI Pre-training with Position-aware Masked Autoencoder}
\author{Kun Wu\footnotemark[1], Zhiguo Jiang\footnotemark[1], Kunming Tang\footnotemark[1],
Jun Shi \footnotemark[2],
Fengying Xie\footnotemark[1],
Wei Wang \footnotemark[3], Haibo Wu \footnotemark[3] \footnotemark[5],
and Yushan Zheng\footnotemark[4] \footnotemark[5]
}

\renewcommand{\thefootnote}{\fnsymbol{footnote}}

\footnotetext[1]{Kun Wu, Zhiguo Jiang, Kunming Tang,  and Fengying Xie are with the Image Processing Center, School of Astronautics, Beihang University, Beijing 100191, China and also with Beijing Advanced Innovation Center on Biomedical Engineering, School of Engineering Medicine, Beihang University, Beijing 100191, China.}

\footnotetext[2]{Jun Shi is with the School of Software, Hefei University of Technology, Hefei 230009, China.}

\footnotetext[3]{Wei Wang and Haibo Wu are with the Department of Pathology, the First Affiliated Hospital of USTC, Division of Life Sciences and Medicine, University of Science and Technology of China, Hefei 230036, China, and also with the Intelligent Pathology Institute, Division of Life Sciences and Medicine, University of Science and Technology of China, Hefei 230036, China (e-mail: wuhaibo@ustc.edu.cn).}

\footnotetext[4]{Yushan Zheng is with Beijing Advanced Innovation Center on Biomedical Engineering, School of Engineering Medicine, Beihang University, Beijing 100191, China (e-mail: yszheng@buaa.edu.cn).}
\footnotetext[5]{\textit{Corresponding author: Yushan Zheng}}

\maketitle
\begin{abstract}
Large-scale pre-training models have promoted the development of histopathology image analysis. However, existing self-supervised methods for histopathology images primarily focus on learning patch features, while there is a notable gap in the availability of pre-training models specifically designed for WSI-level feature learning. In this paper, we propose a novel self-supervised learning framework for pan-cancer WSI-level representation pre-training with the designed position-aware masked autoencoder (PAMA). Meanwhile, we propose the position-aware cross-attention (PACA) module with a kernel reorientation (KRO) strategy and an anchor dropout (AD) mechanism. The KRO strategy can capture the complete semantic structure and eliminate ambiguity in WSIs, and the AD contributes to enhancing the robustness and generalization of the model. We evaluated our method on 7 large-scale datasets from multiple organs for pan-cancer classification tasks. The results have demonstrated the effectiveness and generalization of PAMA in discriminative WSI representation learning and pan-cancer WSI pre-training. The proposed method was also compared with 8 WSI analysis methods. The experimental results have indicated that our proposed PAMA is superior to the state-of-the-art methods. The code and checkpoints are available at https://github.com/WkEEn/PAMA.

\end{abstract}

% \begin{IEEEkeywords}
% WSI pre-training, Pan-cancer, Self-supervised, Transformer, gene mutation
% \end{IEEEkeywords}

\section{Introduction}
\label{sec:introduction}
Digital pathology images have witnessed a significant explosion of whole slide images (WSIs) analysis with deep learning \cite{shmatko2022artificial, dimitriou2019deep
% , korbar2017deep
}. Artificial intelligence framework promotes computer-aided diagnosis for cancer sub-typing \cite{hou2016patch}, histopathology image retrieval \cite{hegde2019similar}, gene mutation prediction \cite{shamai2022deep}, survival prediction \cite{tabibu2019pan, wetstein2022deep}, etc.

Over the past few years, Transformer structures have made impressive gains in the field of natural language processing \cite{vaswani2017attention}. 
Subsequently, many recent studies have further facilitated the WSI analysis by taking advantage of the Transformer to capture and aggregate long-range information \cite{xu2023vision, qian2022transformer, vu2023handcrafted}. 
High-capacity Transformer models have also promoted the development of self-supervised learning \cite{devlin2018bert, atito2021sit}. Self-supervised learning pre-trains a large model on proxy tasks to mine enormous amounts of unlabeled data for potential features and then fine-tunes the model on limited data for specific downstream tasks. The emergence of large-scale models has benefited from the Transformer structure and feature mining of massive data through self-supervised learning, $e.g.$, BERT \cite{devlin2018bert}, CLIP \cite{radford2021learning}, SAM \cite{kirillov2023segment}, and GPT series \cite{openai2023gpt4, nori2023capabilities}. There are an increasing number of studies fine-tuning the pre-trained models on histopathology images, which achieved promising performance in various tasks \cite{lai2023clipath, chauveau2023segment}.

The efficient utilization of unlabeled data firmly fits the trend of pathology image analysis, since there has been an explosion in the volume of pathology image data with the establishment of large open data projects ($e.g.$, the cancer genome atlas program) and the development of online consultation platforms. In this situation, histopathology image foundation models are established based on self-supervised learning frameworks. Typically, Huang \emph{et al} \cite{huang2023visual} applied CLIP \cite{radford2021learning} for
multimodal pathology language-image pre-training (PLIP) based on the public data from medical forums. Similarly, Lu \emph{et al} \cite{lu2024visual} pre-trained a large-scale visual-language foundation model using over 1.17 million image-caption pairs for histopathology analysis. Ikezogwo \emph{et al} \cite{ikezogwo2024quilt} created a multimodal histopathology dataset QUILT-1M within 1M paired image-text samples for CLIP pre-training. 
Nevertheless, the models applied in the above studies were originally designed for natural images and language pre-training. 
Meanwhile, most of the current self-supervised methods for histopathology images focus on learning features of image patches. The substantial resolution of gigapixel WSIs makes it challenging to build an end-to-end framework for WSI-level representation learning. Currently, there is still a lack of available models that can take full advantage of the abundance of histopathology WSIs. 

In this paper, we propose a novel self-supervised learning framework named position-aware masked autoencoder (PAMA) for WSI-level representation learning and pan-cancer pre-training. For the very first time, we propose the slide-level mask image modeling (MIM) proxy task that involves spatial structure to reconstruct WSI representation in feature space. Meanwhile, we embed relative distance and orientation information into slide representation and propose a novel cross-attention module with an orientation dynamic updating strategy and an anchor dropout mechanism.
 We collected 7 large-scale datasets of multiple organs to evaluate the effectiveness and generalization of our proposed framework with slide-level representation learning and multi-organ pre-training and compared it with 8 SOTA WSI analysis methods. The experimental results have demonstrated that PAMA is effective in histopathology WSI pre-training and downstream tasks, including cancer sub-typing and biomarkers prediction.

We summarize the contribution of the paper in three aspects.
\begin{enumerate}
    \item We propose a novel self-supervised learning framework based on the position-aware masked autoencoder named PAMA for WSI pre-training. We train PAMA on the slide-level MIM proxy task to reconstruct WSI representation in the feature space which can sufficiently mine the semantic features of histopathology slides from a large amount of unlabeled data.
    \item We propose the anchor-based position-aware cross-attention (PACA) module to enable bidirectional communication between the local and global information of WSIs. An anchor dropout mechanism is introduced for augmentation to facilitate the robustness and generalization of PAMA. Meanwhile, the relative distance and orientation information are embedded into slide features to maintain comprehensive spatial semantics. Additionally, we introduce a kernel reorientation (KRO) strategy to dynamically update the main orientation of anchors for better obtaining complete semantic structure and eliminating ambiguity.
    \item We evaluated the proposed method on 7 large-scale datasets containing 13,685 WSIs from multiple organs for multiple diagnostic tasks. The results demonstrate that pan-cancer pre-training facilitates PAMA's significant progress in fine-grained WSI-level tasks, including biomarkers prediction and cancer sub-typing. Furthermore, PAMA achieves the best performance over the other 8 SOTA methods.
\end{enumerate}
A previous version of the paper has been published in the conference paper \cite{wu2023position}.
\begin{figure*}[t]
\centering
\includegraphics[width=\linewidth]{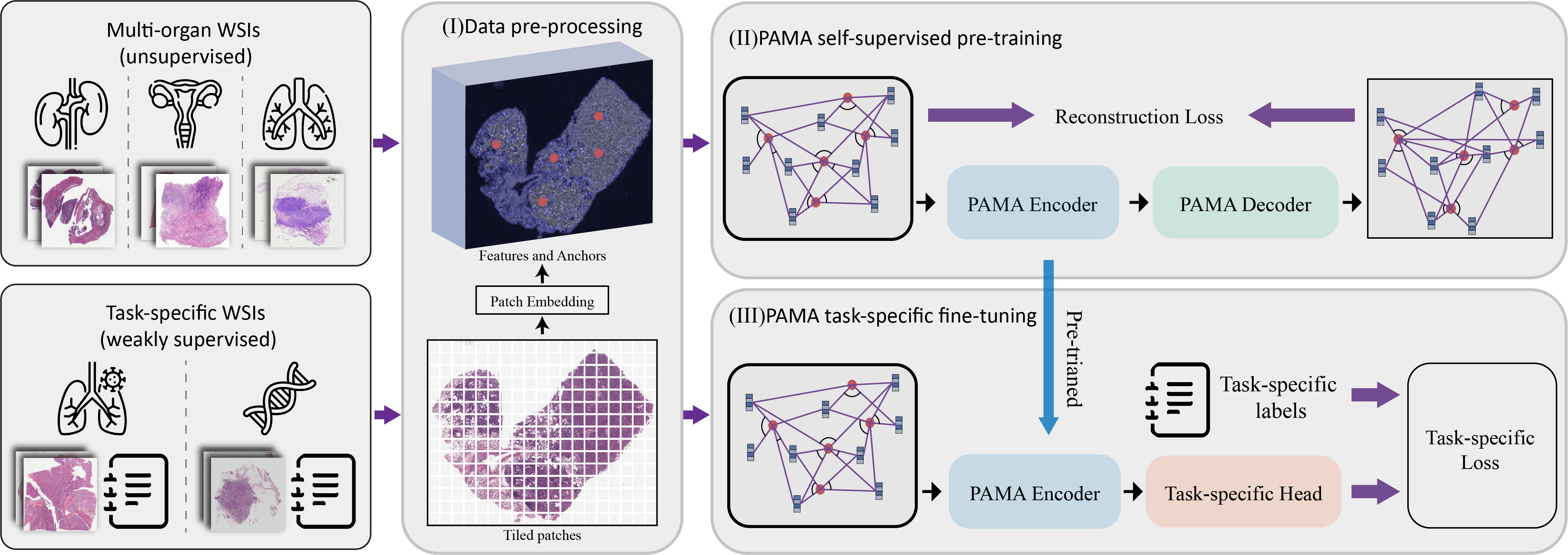}
\caption{Framework of pan-cancer WSI pre-training and task-specific fine-tuning, where (I) is the data pre-processing in which the spatial and structural information are constructed for the WSI feature, (II) displays 
 the pre-training stage based on reconstructing the slide representation on label-free multi-organ datasets, and (III) is the fine-tuning of the encoder on weakly-supervised task-specific data for practical inference. }
\label{framework}
\end{figure*}

\section{Related works}

\subsection{MIL based methods}
The WSI is gigapixel large-scale image data which makes it challenging to apply the end-to-end deep learning framework to analyze WSIs as in the case of natural scene images. Two-stage methods are generally employed in slide analysis, involving the extraction of patch features and aggregation of WSI-level representation.

Multiple instance learning (MIL) has become the typical solution for slide representation aggregation \cite{campanella2019clinical}. For instance, Li \emph{et al} \cite{li2021dual} proposed a dual-stream framework to integrate the instances and applied a pyramidal fusion mechanism for multiscale WSI features.
Some studies have introduced new techniques into the aggregation stage to describe the spatial structure of WSI.
Graph Attention MIL \cite{raju2020graph} and LAGE-Net \cite{zheng2022encoding} constructed the graph structure of patches to encode local relationships. However, the methods were difficult to capture long-range spatial information.
Thereby, Transformer methods based on the self-attention mechanism are introduced into MIL to aggregate the global features of WSIs.
The Transformer structure is adapted to gather long-range features and comprehend overall structural connections, making it suitable for large-scale WSI analysis. TransMIL \cite{shao2021transmil} and SETMIL \cite{zhao2022setmil} leveraged some CNN blocks and spatial encoding modules to aggregate local information and used the self-attention model for global messaging. 
These methods disregarded the isotropic characteristics of pathology images, potentially resulting in ambiguous position encoding. To address this problem, KAT \cite{zheng2023kernel} constructed hierarchical masks based on local kernels to preserve multi-scale relative distance information in training. However, these masks were manually specified, which means they are not trainable and lack dynamic orientation information.

\subsection{Self-supervised learning}
Self-supervised learning methods have gained considerable interest in computer vision, frequently concentrating on diverse proxy tasks for pre-training without any manual annotations \cite{luo2022self, an2022masked, shi2022semi, wang2022transformer}. The label-free approaches facilitate patch representation learning to release resource consumption from fine-grained annotation. Some works focused on context-based methods, such as predicting pathology image cross-stain, predicting the resolution sequence ordering in WSI, and constructing associations between proximity and feature similarity \cite{gildenblat2019self, yang2021self, srinidhi2022self}. Other methods leveraged generative models to build proxy tasks that implicitly learn features by minimizing the reconstruction loss in the pixel space, like SD-MAE \cite{luo2022self} and MAE-MIL \cite{an2022masked}. More approaches applied contrastive learning to enhance patch feature learning. CTransPath \cite{wang2022transformer} proposed a semantically relevant contrastive learning framework that compares relevance between instances to mine more positive pairs. TransPath \cite{wang2021transpath} used BYOL \cite{grill2020bootstrap} architecture due to its negative sample independence and proposed a token-aggregating and excitation (TAE) module for capturing more global information. 
Chen \emph{et al} used DINOv2 \cite{oquab2023dinov2}, a state-of-the-art self-supervised learning method based on student–teacher knowledge distillation for pre-training large ViT architectures, for large-scale visual pre-training on 100,426 histology slides. Vorontsov \emph{et al} \cite{vorontsov2024foundation} presented a million-image-scale pathology foundation model, Virchow, pre-trained on data from approximately 100,000 patients corresponding to approximately 1.5 million WSIs. There are also many studies introduce multimodal data into self-supervised pre-training for histopathology image representation learning \cite{lu2024visual, ikezogwo2024quilt}.
However, these patch-level representation learning methods treated patches as independent entities, thereby destroying the integrity of the semantic information in the WSI. Furthermore, under conditions of limited annotation information, such an approach would yield over-fit the slide-level aggregation model.

HIPT \cite{chen2022scaling} investigated the novel concept of slide-level self-supervised learning, representing a significant challenge. Chen \emph{et al} \cite{chen2022scaling} constructed a two-stage self-supervised framework in which DINO \cite{caron2021emerging} is utilized to pre-train patches (256×256) and then another DINO is pre-trained for the regions (4096×4096) of WSIs. HIPT leveraged the hierarchical structure inherent in WSIs to construct a multi-level self-supervised learning framework. By doing this, the framework learned high-resolution image representations, enabling it to benefit from the plentiful unlabeled WSIs. This contributes to an increase in the accuracy and robustness of tumor recognition. 
Recently, Xu \emph{et al} \cite{xu2024whole} proposed Prov-GigaPath, a slide-level representation leaning framework pre-trained on 171,189 slides originated from more than 30,000 patients covering 31 major tissue types.
Nevertheless, the ViT backbone employed for HIPT ignores the structural characteristics of large-scale pathology images. Additionally, the multi-stage pre-training may result in the accumulation of bias and error, reducing the performance of the final model. 
Prov-GigaPath applies LongNet \cite{ding2023longnet} as a slide aggregator, leveraging its design for extremely long sequences. However, Prov-GigaPath does not account for the unique characteristics of WSIs, particularly their spatial structure.

\begin{figure*}[t]
\centering
\includegraphics[width=\linewidth]{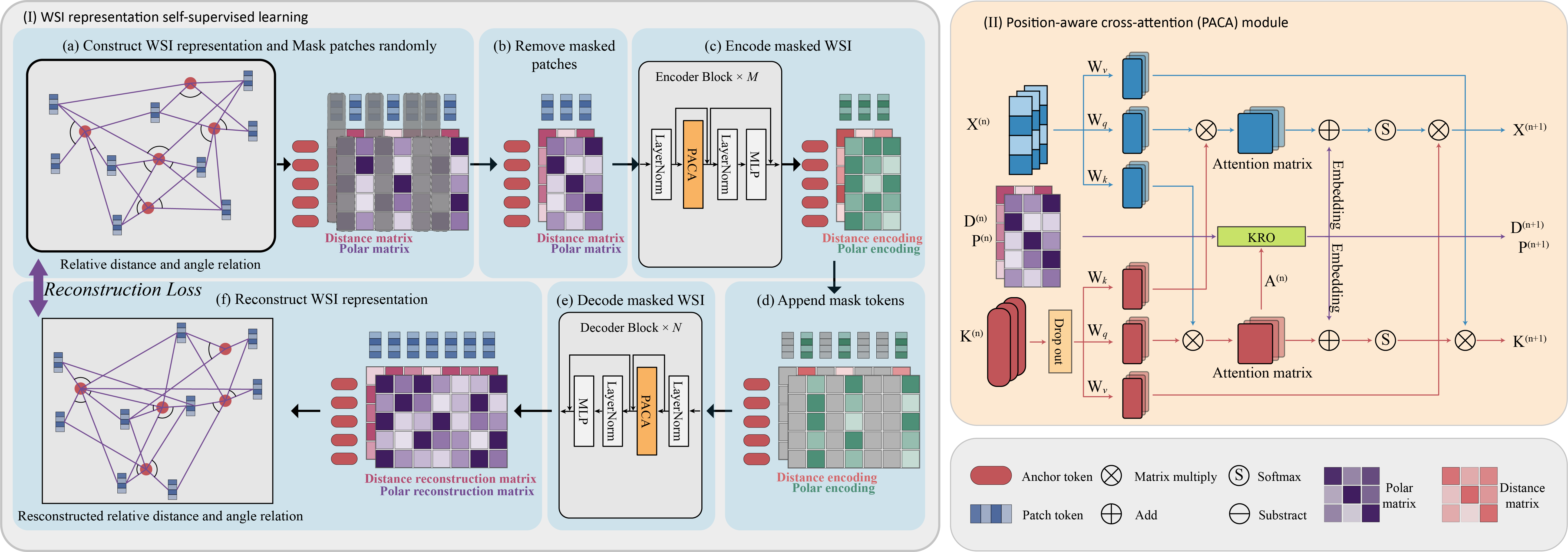}
\caption{Illustrations of each structure in PAMA, where (I) describes the workflow of WSI representation self-supervised learning with PAMA, including encoder, decoder, and slide representation reconstruction, (II) is the structure of the position-aware cross-attention (PACA) module which is the core of PAMA, in which the kernel reorientation (KRO) strategy is described in Algorithm ~\ref{algorithm1} and the detailed process of anchor dropout is described in section ~\ref{AD}.}
\label{subs}
\end{figure*}

\subsection{Pan-cancer analysis}
There is inter-patient heterogeneity across different types of cancer which means tumors of different cancer types may share underlying similarities \cite{cheerla2019deep}. Therefore, pan-cancer analysis of large-scale data across a broad range of cancers can potentially improve disease modeling by exploiting these pan-cancer similarities. A growing number of works are focusing on building pan-cancer analytical models and related databases through computational pathology. Komura \emph{et al} \cite{komura2022universal} built a universal encoder for cancer histology through a deep neural network. It allows for genomic feature prediction from histology images across various cancer types. Yu \emph{et al} \cite{fu2020pan} employed deep transfer learning to quantify histopathological patterns in 17,355 WSIs from 28 cancer types. Subsequently, they correlated these patterns with matched genomic, transcriptomic, and survival data. PanNuke \cite{gamper2019pannuke} is an open pan-cancer histology dataset for nuclei instance segmentation and classification across 19 different tissue types. However, there is still a lack of a pan-cancer analysis model that can utilize a large number of unlabeled WSIs for slide-level feature learning.

\section{Methods}
\subsection{Overview}
We propose the position-aware masked autoencoder (PAMA) following the self-supervised learning protocol shown in Fig. \ref{framework}. 
After data pre-processing, we construct spatial and structural information of WSIs for their slide-level representation. Our proposed PAMA encodes the slide representation into a latent space and then decodes the latent feature back to the origin feature space for reconstructing the slide-level representation. The proxy task of reconstructing position-aware slide-level representation trains the PAMA to capture complicated semantics and improve generalization. 

\subsection{Positon-aware Masked autoencoder (PAMA)}
\subsubsection{Problem formulation}
MAE \cite{he2022masked} is a promising paradigm for image representation learning. We introduce masked autoencoder into histopathology slide-level representation learning. Unlike natural scene images, histopathology digital images are scale-varying and semantically complex which is challenging to capture the complete semantic structure and eliminate ambiguity.
To combat the limitation, we propose a position-aware structure to construct slide representation. Firstly, patch features of a slide are extracted, which is formulated as $\textbf{X} \in \mathbb{R}^{n_p \times d_f}$, where $n_p$ is the number of patches in the slide and $d_f$ is the dimension of the patch feature.
Inspired by the way to describe spatial information in KAT \cite{zheng2023kernel}, multiple anchors are selected by clustering the location coordinates of all patches for profiling local structural semantics. The learnable vectors are assigned for these anchors formulated as $\textbf{K} \in \mathbb{R}^{n_k \times d_f}$, where $n_k = [\frac{n_p}{c}]$ is the number of anchors in the slide with $c$ representing the expected number of patches per cluster. Additionally, a polar coordinate system is constructed where each anchor is regarded as a pole. In this system, every patch has explicit relative distance and angle definitions to each anchor. Therefore, we define $\textbf{D} \in \mathbb{N}^{n_k \times n_p}$ and $\textbf{P} \in \mathbb{N}^{n_k \times n_p}$ to represent the relative distance matrix and the relative polar angle matrix of a WSI, respectively. We equate the polar and distance values into bins to ensure adaptation to scale-varying slides, so the inputs are discrete integers, similar to the positional index of each token in Transformer\cite{vaswani2017attention}.
Specifically, ${D_{ij}} \in \textbf{D}$ and ${P_{ij}} \in \textbf{P}$ correspondingly denote the relative distance and relative polar angle of the $j$-th patch to the $i$-th anchor. Based on the above, a WSI is formulated as $S = \{ \textbf{X},\textbf{K},\textbf{D},\textbf{P} \}$.
We leverage the anchor-based and position-aware data structure to represent a WSI, which can adaptively maintain the spatial integrity and semantic enrichment of scale-varying slides across multiple organs.
\subsubsection{PAMA encoder}
Our proposed position-aware masked autoencoder is shown in Fig.\ref{subs}(I). Some patch tokens of the original WSI feature are randomly masked with a high ratio ($i.e.$, ratio $r=75\%$) and these tokens including their corresponding spatial information are removed.
The remaining ($i.e.$, unmasked) tokens are fed into our encoder. 
Each encoder block is shown in Fig. \ref{subs}(I)(c), which is formulated as follows
\begin{align}
\textbf{$\hat{\textbf{X}}$}_{l}, \textbf{$\hat{\textbf{K}}$}_{l} &= LayerNorm([\textbf{X}_{l-1}; \textbf{K}_{l-1} ]),  \\
\textbf{$\Tilde{\textbf{X}}$}_{l}, \textbf{$\Tilde{\textbf{K}}$}_{l} &= PACA([\textbf{$\hat{\textbf{X}}$}_{l}; \textbf{$\hat{\textbf{K}}$}_{l}; \textbf{P}_{l}; \textbf{D}_{l}]), \\
\textbf{X}_{l} &= \textbf{$\Tilde{\textbf{X}}$}_{l} + MLP(LayerNorm(\textbf{$\Tilde{\textbf{X}}$}_{l} + \textbf{$\hat{\textbf{X}}$}_{l})), \\
\textbf{K}_{l} &= \textbf{$\Tilde{\textbf{K}}$}_{l} + MLP(LayerNorm(\textbf{$\Tilde{\textbf{K}}$}_{l} + \textbf{$\hat{\textbf{K}}$}_{l})), 
\end{align}
\label{eq_1}
where $MLP$ denotes multilayer perception, $PACA$ is our proposed position-aware cross-attention module which will be detailed later, and $l$ is the index of the block. Our encoder maps the sparse WSI features into a latent representation and meanwhile maintains the spatial information.

\subsubsection{PAMA decoder}
We adopt an asymmetric design in the decoder. The input to the decoder is a complete set of tokens, consisting of $\textbf{X}$ and the masked tokens $\textbf{M} \in \mathbb{R}^{(n_p \times r) \times d_f}$ as shown in Fig.~\ref{subs}(I)(d). The $\textbf{M}$ are initialized with trainable vectors and the corresponding spatial embeddings are added. 
For reconstructing the WSI representation, we decode $\{\textbf{X};\textbf{M}\}$ into the original feature space and calculate the loss only on the masked tokens between the reconstructed and original features as shown in Fig.~\ref{subs}(I)(f).
The proxy task to predict masked tokens based on the sparse WSI feature can assist our PAMA in acquiring adaptive WSI-level representation while guaranteeing the integrity of spatial information and pathology semantics.

\subsubsection{Objectives}
Referring to the MAE \cite{he2022masked} structure, we append a $\textbf{X}_{class}$ token before all patch tokens to represent the learned slide feature and feed the $\textbf{X}_{class}$ token into the task-specific head for inference. In the pre-training phase, the $\textbf{X}_{class}$ token does not participate in loss computation, but it consistently communicates with anchors and gathers global information. Subsequently, the pre-trained parameters of the $\textbf{X}_{class}$ token will be used for fine-tuning and linear probing. Finally, we calculate the mean squared error (MSE) on the masked tokens between the reconstructed and original features.

\subsection{Position-aware cross-attention (PACA)}
We propose the position-aware cross-attention (PACA) module to build bidirectional message passing between anchors and patches. Fig. \ref{subs} (II) illustrates the structure of PACA. From the perspective of anchors, different local regions should respond dynamically to all patches as below:
\begin{align}
&\textbf{A}^{(n)} = \sigma(\frac{{\textbf{$\hat{\textbf{K}}$}^{(n)}\textbf{W}_q^{(n)}}\cdot {(\textbf{$\hat{\textbf{X}}$}^{(n)}\textbf{W}_k^{(n)})}^{\mathrm{T}}}{\sqrt{{d_e}}} +\varphi_{d} (\textbf{D}^{(n)})+\varphi_{p} (\textbf{P}^{(n)})), \\
&\textbf{$\hat{\textbf{K}}$}^{(n+1)} = \textbf{A}^{(n)} \cdot (\textbf{$\hat{\textbf{X}}$}^{(n)}\textbf{W}_v^{(n)}),
\end{align}
\label{eq2} 
% \end{equation}
where $\textbf{W}_{q,k,v} \in \mathbb{R}^{d_f \times d_e}$ are trainable parameters and $d_e$ denotes the dimension of the head output, $\varphi_{d}$ and $\varphi_{p}$ are the transformation functions that respectively map the distance and polar angle to corresponding learnable embedding values, $\sigma$ is the softmax function and $n$ is the index of layer. We apply two transformation functions to embed polar and distance into vectors, respectively, to ensure that the positional information is continuous and trainable. We add position embeddings as bias in softmax function to effectively facilitate the module to capture global information, drawing inspiration from the Graphormer \cite{ying2021transformers}.

Symmetrically, each patch token updates its representation by catching the local region information from all anchors as below:

\begin{align}
&\textbf{$\bar{\textbf{A}}$}^{(n)} = \sigma(\frac{{\textbf{$\hat{\textbf{X}}$}^{(n)}\textbf{W}_q^{(n)}}\cdot {(\textbf{$\hat{\textbf{K}}$}^{(n)}\textbf{W}_k^{(n)})}^{\mathrm{T}}}{\sqrt{{d_e}}} +\varphi_{d}^{\mathrm{T}} (\textbf{D}^{(n)})+\varphi_{p}^{\mathrm{T}} (\textbf{P}^{(n)})), \\
&\textbf{$\hat{\textbf{X}}$}^{(n)} = \textbf{$\bar{\textbf{A}}$}^{(n)} \cdot (\textbf{$\hat{\textbf{K}}$}^{(n)}\textbf{W}_v^{(n)}),
\end{align}
\label{eq3}

The transmission of local information and perception of global information occurs promptly due to the two-way communication between patches and anchors. The model maintains the semantic and structural integrity of the WSI and prevents representation collapse in the local area throughout the training process with the embedding of relative distance and polar angle information. Regarding efficiency, the computational complexity of self-attention is $O({n_p}^2)$, where $n_p$ represents the number of patch tokens. Conversely, our proposed PACA has a complexity of $O(n_k × n_p)$, where $n_k$ represents the number of anchors. It is important to note that when $n_k << n_p$, the complexity is nearly $O(n_p)$, which exhibits a linear correlation with the WSI's size.

\begin{algorithm}[t]\scriptsize
    \label{algorithm1}
    \SetAlgoLined
    \KwIn{
    \\${\textbf{P}^{(n)}\in \mathbb{N}^{H \times \hat{n}_k \times n_p}}$: The relative polar angle matrix of $n$-th block, where $H$ is the head number of multi-head attention, $n_p$ is the number of patches in the WSI, $ \hat{n}_k = n_k \times p$ where $n_k$is the number of anchors in the WSI and $p$ is the probability of anchor dropout;
    \\${\textbf{A}^{(n)}\in \mathbb{R}^{H \times \hat{n}_k \times n_p}}$: The attention matrix from anchors to patches, defined as 
    $\textbf{A}^{(n)} =\sigma(\frac{{\hat{\textbf{K}}^{(n)}\textbf{W}_q^{(n)}}\cdot {(\hat{\textbf{X}}^{(n)}\textbf{W}_k^{(n)})}^{\mathrm{T}}}{\sqrt{{d_e}}} +\varphi_{d} (\textbf{D}^{(n)})+\varphi_{p} (\textbf{P}^{(n)}))$
    \\${D^{score}}$: A dictionary taking the angle as KEY for storing attention scores;
    \\$N$: The number of orientation bins assigned to each anchor;}
    \KwOut{${\textbf{P}^{(n+1)}\in \mathbb{R}^{H \times \hat{n}_k \times n_p}}$: The updated polar angle matrix.}
    \For{$h$ in $H$}{
        \For{$i$ in $\hat{n}_k$}{
            Initialize $D^{score}$ with $\mathbf{0}$\\
            \For{$j$ in $n_p$}{
                $D^{score}[\textbf{P}^{(n)}_{h,i,j}]$ += $\textbf{A}^{(n)}_{h,i,j}$;
            }
            $\textbf{P}^{(n)}_{h,i,max}=\arg\max D^{score}$; \tcp{Find the orientation that has the highest attention score.} 
            \For{$j$ in $n_p$}{
                $\textbf{P}^{(n+1)}_{h,i,j} = (\textbf{P}^{(n)}_{h,i,j} - \textbf{P}^{(n)}_{h,i,max}) \ mod \ N$; \tcp{Reorientation.}
            }
        }
    }
    \caption{Kernel Reorientation algorithm.} 
\end{algorithm}     

\begin{figure}[!t]
	\centering
	\includegraphics[width=\linewidth]{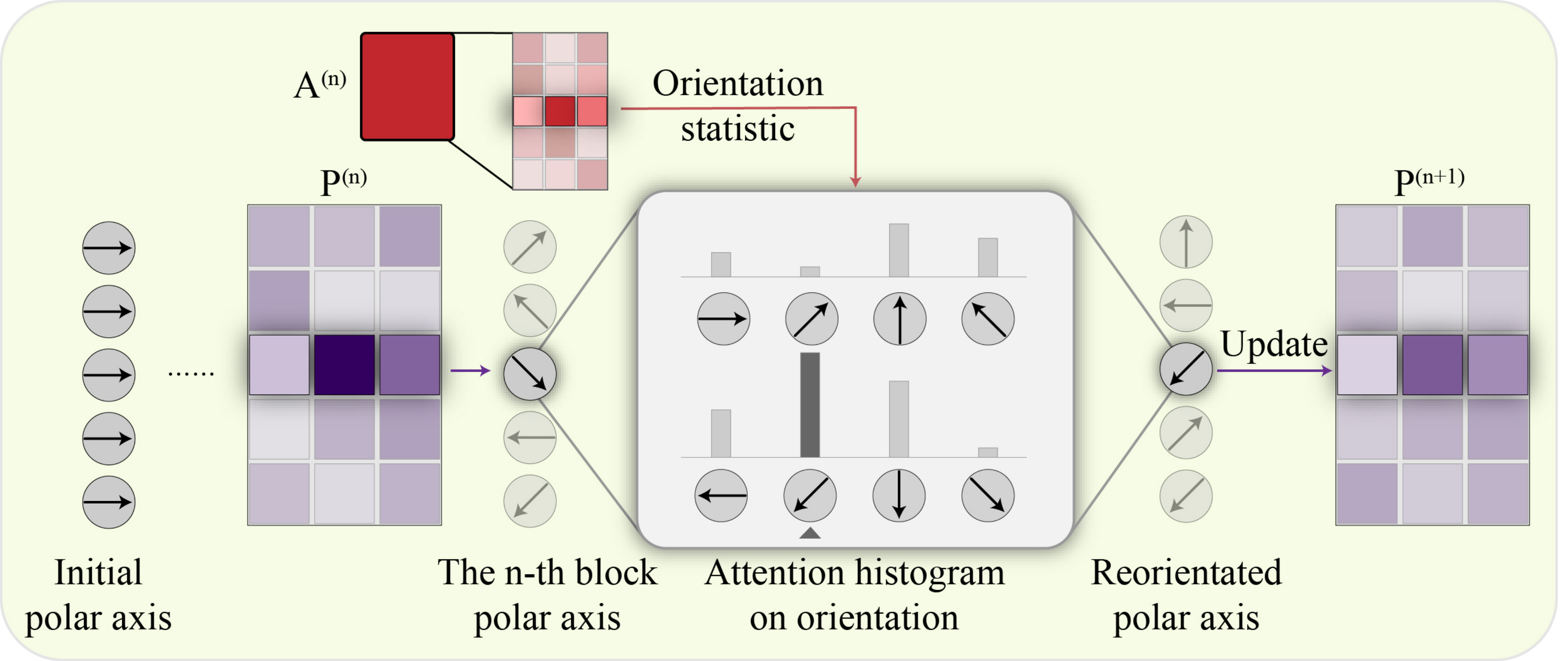}
	\caption{The illustration of the proposed Kernel Reorientation (KRO) strategy, where we show the KRO process for an anchor and highlight it with shading effects for secretarial clarity.}
	\label{fig:kro}
\end{figure}

\subsection{Kernel Reorientation (KRO)}
In natural scene images, there is a directional conspicuousness of semantics. For example, in a church, it is more common for the door to be positioned below the windows rather than above. However, histopathology images do not have an absolute definition of the main direction. The meaning of a WSI remains invariant under rotation and flipping. Namely, it is isotropic. Embedding orientation information with a fixed polar axis will result in ambiguities in multiple slides. Therefore, we propose the kernel reorientation (KRO) strategy to dynamically update every anchor's main polar axis.

As shown in Fig \ref{fig:kro}, we illustrate the KRO strategy in detail. Regarding the polar angle matrix $\textbf{P}^{(n)} \in \mathbb{N}^{n_k \times n_p}$ during the \textit{n}-th block, the initial polar axis is defined as the horizontal direction for all the anchors. For each anchor, the orientation is divided into $N$ equal bins. For example, if $N=8$, each bin corresponds to a $\frac{\pi}{4}$ sector. During the processing of PACA, an attention score matrix of all anchors to patches formulated as $\textbf{A}^{(n)} \in \mathbb{R}^{n_k \times n_p}$ is obtained which reflects the contribution from patches to anchors. Based on the matrix $\textbf{A}^{(n)}$, we calculate the attention histogram on the orientation for each anchor by summarizing the attention score of all patches within each orientation bin. Then, the bin with the max score is selected as the new polar main axis, $i.e.$, the reorientated polar axis. Based on the new axis of anchors, we update the polar angle of patches and obtain the updated matrix $\textbf{P}^{(n+1)}$. The detailed algorithm is outlined in Algorithm \ref{algorithm1}.

\subsection{Anchor Dropout (AD)}
\label{AD}
We defined anchors following the over-saturated strategy, which is similar to neurons in the neural network. The anchors are clustered based on spatial coordinates, which are proxies of local region information. Fixing the anchor position of the WSI across training epochs will result in losing the flexibility of local relationships and redundancy. Inspired by the neurons dropout mechanism \cite{srivastava2014dropout}, we introduce anchor dropout to enhance the robustness and generalization of the model and relieve the over-fitting in the WSI pre-training. The dropout is applied before Eq.~\ref{eq_1} with the following equations
\begin{align}
b_l^{(k)} &\sim Bernoulli(p), k =0,...,n_k-1, \\
\textbf{K}_{l} &:= Index(\textbf{K}_{l},\textbf{b}_l) , \textbf{b}_l = [b_l^{(0)},...,b_l^{(n_k-1)}],
\end{align}
where $p$ is the probability of dropout, $\textbf{b}_l$ is a vector of independent Bernoulli random variables each of which has a probability $p$ of being 1, and $ Index(\textbf{K}_{l},\textbf{b}_l) $ means returning a subset of $\textbf{K}_{l}$ based on the corresponding index in $\textbf{b}_{l}$. 
% The random dropout of anchors facilitates the capture of more diverse and rich local representations for each WSI. 

\section{Experiments}
\subsection{Datasets}
We collected four public large-scale datasets from the cancer genome atlas (TCGA) program and three in-house datasets to evaluate our method, which are introduced as follows:
\begin{itemize}
    \item[$\bullet$] \textbf{TCGA-RCC} contains 659 WSIs of renal cell carcinoma (RCC) patients, which are categorized into 3 subtypes including kidney renal clear cell carcinoma (KIRC), kidney renal papillary cell carcinoma (KIRP), and kidney renal chromophobe cell carcinoma (KICH).
    \item[$\bullet$] \textbf{TCGA-NSCLC} contains 3,064 WSIs of non-small cell lung cancer (NSCLC) patients from the TCGA program, which are categorized into 3 subtypes including tumor-free (Normal), lung adenocarcinoma (LUAD), and lung squamous cancer (LUSC).
    \item[$\bullet$] \textbf{USTC-EGFR} contains 531 in-house WSIs of lung adenocarcinoma for epidermal growth factor receptor (EGFR) gene mutation identification, which are categorized into 4 subtypes including EGFR 19del mutation, EGFR L858R mutation, Non-common driver mutations (Wild type), and other driver gene mutations.
    \item[$\bullet$] \textbf{Endometrium-3k} contains 3,654 in-house WSIs of endometrial pathology including 8 categories, namely well/moderately/low-differentiated endometrioid adenocarcinoma  (WDEA/MDEA/LDEA), squamous differentiation endometrioid carcinoma (SDEC), plasmacytoid endometrioid carcinoma (PECA), clear cell endometrioid carcinoma (CCEC), mixed-cell endometrioid adenocarcinoma (MCEA), and tumor-free (Normal). 
    \item[$\bullet$] \textbf{TCGA-EGFR} contains 705 WSIs of lung adenocarcinoma with EGFR gene mutation, which are categorized into 2 subtypes including EGFR mutation and Wild type.
    \item[$\bullet$] \textbf{BRCA-HER2} contains 279 in-house WSIs of human epidermal growth factor receptor-2 (HER2) protein and gene expression in breast cancer patients, which are categorized into 4 subtypes including the IHC score of 1+, the IHC score of 2+, the IHC score of 3+, and the IHC score of 0 (Normal).
    \item[$\bullet$] \textbf{TCGA pan-cancer dataset} contains 4,793 unlabeled WSIs containing 10 cancer types from 7 primary sites as shown in Table \ref{table_tcgaPan}, which is collected from TCGA program designedly for evaluation of generalization for out-of-domain pre-training. 
\end{itemize}
These datasets, except for TCGA pan-cancer dataset, consist of 8,892 WSIs from multiple organs and can be used for studies such as cancer sub-typing, molecular status prediction, and gene mutation prediction, which are all slide-level tasks. We randomly divided every dataset into training, validation, and testing subsets according to the ratio of 6:1:3, where the training sets were used for multi-organ pre-training and task-specific fine-tuning, validation sets were used to do early stop, and results on the testing sets were reported for evaluation. We describe the task definitions on these datasets and the utilization of the data under the multi-organ pre-training strategy as shown in Table \ref{table_statisticData}, where tasks are categorized into in-domain and out-of-domain conditions based on whether or not the fine-tuning data is involved in the pre-training process.

\begin{table}[h]
\centering
\arrayrulecolor{black}
{
\caption{Detailed data distribution of TCGA pan-cancer dataset.}
\label{table_tcgaPan}
\small
\setlength{\tabcolsep}{5pt}
\begin{tabular}{l|l|c}
\hline
\textbf{Dataset} & \textbf{Primary Site} & \textbf{Number of WSIs} \\ \hline \hline
TCGA-BRCA & Breast           & 1121 \\ \hline
TCGA-CESC & Gynecology       & 278  \\ \hline
TCGA-BLCA & Urinary          & 457  \\ \hline
TCGA-PAAD & Liver            & 205  \\ \hline
TCGA-COAD & \multirow{3}{*}{Gastrointestinal} & 441  \\ 
TCGA-READ &  & 158  \\ 
TCGA-STAD &  & 400  \\ \hline
TCGA-PRAD & Prostate         & 449  \\ \hline
TCGA-GBM  & Brain            & 816  \\ \hline
TCGA-HNSC & Head and Neck    & 468  \\ \hline
\end{tabular}
}
\end{table}

\begin{table*}[!t]
\centering
\arrayrulecolor{black}
{
\caption{Definition and data distribution table for all tasks under the multi-organ pre-train strategy, where we define two conditions, In-domain and Out-of-domain, based on whether the fine-tune dataset participates in pre-train or not.}
\label{table_statisticData}
\fontsize{5.8}{10}\selectfont
\setlength{\tabcolsep}{1.7pt}
\begin{tabular}{l|ccccc|cccccc}
\hline
\textbf{Pre-training condition} & \multicolumn{5}{c|}{\textbf{In-domain}} & \multicolumn{6}{c}{\textbf{Out-of-domain}} \\ \hline
\textbf{Pre-training data} & \multicolumn{5}{c|}{\makecell[c]{Training subsets of Endometrium-3k, TCGA-NSCLC, \\TCGA-RCC,USTC-EGFR, and TCGA-EGFR}} & \multicolumn{6}{c}{TCGA pan-cancer dataset} \\ \hline \hline
\multicolumn{12}{c}{\textbf{Downstream Tasks}} \\ \hline \hline
\textbf{Organ} &  Endometrium & Lung & Kidney & Lung & Lung & Endometrium & Lung & Kidney & Lung & Lung & Breast \\ \hline
\multirow{2}{*}{\textbf{Fine-tuning data}}  & \multicolumn{5}{c|}{\textbf{Training subset of}} & \multicolumn{6}{c}{\textbf{Training subset of}} \\
 & Endometrium-3k & TCGA-NSCLC & TCGA-RCC & USTC-EGFR & TCGA-EGFR & Endometrium-3k & TCGA-NSCLC & TCGA-RCC & USTC-EGFR & TCGA-EGFR & BRCA-HER2 \\ \hline
\multirow{2}{*}{\textbf{Evaluation data}} & \multicolumn{5}{c|}{\textbf{Testing subset of}} & \multicolumn{6}{c}{\textbf{Testing subset of}} \\
& Endometrium-3k & TCGA-NSCLC & TCGA-RCC & USTC-EGFR & TCGA-EGFR & Endometrium-3k & TCGA-NSCLC & TCGA-RCC & USTC-EGFR & TCGA-EGFR & BRCA-HER2 \\ \hline
\end{tabular}
}
\end{table*}

\begin{table*}[!t]
\centering
\scriptsize
\caption{Results of fine-tuning on multi-organ datasets with diffident training strategies under DINO \cite{caron2021emerging} patch features and PLIP\cite{huang2023visual} patch features, where the results of \colorbox{cyan!20}{pre-trained on multi-organ} are in the cyan background and results of \colorbox{lightgray!50}{pre-trained on single-organ} are in the gray background.}
\label{table1}
\setlength{\tabcolsep}{5pt}
\begin{tabular}{l|l|lll|lll}
\hline
\multirow{2}{*}{Datasets}& \multirow{2}{*}{Strategies}& \multicolumn{3}{c|}{DINO \cite{caron2021emerging}} & \multicolumn{3}{c}{PLIP \cite{huang2023visual}}\\
& & ACC (\%)& AUC & F1 score & ACC (\%)& AUC & F1 score \\
\hline 
\hline
\multirow{3}{*}{\makecell[l]{Endometrium-3k}}
& w/o pre-train & 38.67 & 0.837 & 0.424 & 38.43 & 0.801 & 0.384 \\
& \cellcolor{lightgray!50}single-organ 
& \cellcolor{lightgray!50}{47.47(+22.75\%)} & \cellcolor{lightgray!50}{0.855(+2.15\%)} & \cellcolor{lightgray!50}{0.464(+9.43\%)} 
& \cellcolor{lightgray!50}{42.16(+9.71\%)} & \cellcolor{lightgray!50}{0.837(+4.49\%)} & \cellcolor{lightgray!50}{0.422(+9.89\%)} \\
& \cellcolor{cyan!20}multi-organ 
& \cellcolor{cyan!20}{50.12(+29.61\%)} & \cellcolor{cyan!20}{0.877(+4.77\%)} & \cellcolor{cyan!20}{0.483(+13.90\%)} 
& \cellcolor{cyan!20}{45.06(+17.25\%)} & \cellcolor{cyan!20}{0.883(+10.23\%)} & \cellcolor{cyan!20}{0.451(+17.44\%)}\\ 
  \hline
\multirow{3}{*}{\makecell[l]{TCGA-NSCLC}}
& w/o pre-train & 86.19 & 0.971 & 0.884 & 86.43 & 0.967 & 0.862 \\
& \cellcolor{lightgray!50}single-organ 
& \cellcolor{lightgray!50}{92.72(+7.57\%)} & \cellcolor{lightgray!50}{0.988(+1.75\%)} & \cellcolor{lightgray!50}{0.919(+3.95\%)} 
& \cellcolor{lightgray!50}{87.28(+0.98\%)} & \cellcolor{lightgray!50}{0.971(+0.41\%)} & \cellcolor{lightgray!50}{0.865(+0.35\%)}\\
& \cellcolor{cyan!20}multi-organ 
& \cellcolor{cyan!20}{93.51(+8.49\%)} & \cellcolor{cyan!20}{0.989(+1.85\%)} & \cellcolor{cyan!20}{0.924(+4.52\%)} 
& \cellcolor{cyan!20}{87.61(+1.36\%)} & \cellcolor{cyan!20}{0.976(+0.93\%)} & \cellcolor{cyan!20}{0.876(+1.62\%)} \\ 
 \hline
\multirow{3}{*}{\makecell[l]{TCGA-RCC}}
& w/o pre-train & 91.72 & 0.978 & 0.914 & 85.25 & 0.976 & 0.853 \\
& \cellcolor{lightgray!50}single-organ
& \cellcolor{lightgray!50}{91.88(+0.17\%)} & \cellcolor{lightgray!50}{0.981(+0.31\%)} & \cellcolor{lightgray!50}{0.917(+0.33\%)} 
& \cellcolor{lightgray!50}{90.64(+6.32\%)} & \cellcolor{lightgray!50}{0.987(+1.12\%)} & \cellcolor{lightgray!50}{0.908(+6.44\%)}\\
& \cellcolor{cyan!20}multi-organ
& \cellcolor{cyan!20}{92.46(+0.81\%)} & \cellcolor{cyan!20}{0.989(+1.12\%)} & \cellcolor{cyan!20}{0.925(+1.20\%)} 
& \cellcolor{cyan!20}{93.88(+10.12\%)} & \cellcolor{cyan!20}{0.991(+1.53\%)} & \cellcolor{cyan!20}{0.939(+10.08\%)} \\ 
 % \cline{2-7}
 \hline
\multirow{3}{*}{\makecell[l]{USTC-EGFR}} 
& w/o pre-train & 83.03 & 0.804 & 0.494 & 83.63 & 0.813 & 0.509\\
& \cellcolor{lightgray!50}single-organ 
& \cellcolor{lightgray!50}{86.27(+3.90\%)} & \cellcolor{lightgray!50}{0.807(+0.37\%)} & \cellcolor{lightgray!50}{0.528(+6.88\%)} 
& \cellcolor{lightgray!50}{85.45(+2.18\%)} & \cellcolor{lightgray!50}{0.828(+1.85\%)} & \cellcolor{lightgray!50}{0.552(+8.45\%)} \\
& \cellcolor{cyan!20}multi-organ 
& \cellcolor{cyan!20}{87.88(+5.84\%)} & \cellcolor{cyan!20}{0.826(+2.74\%)} & \cellcolor{cyan!20}{0.572(+15.78\%)} 
& \cellcolor{cyan!20}{86.16(+3.03\%)} & \cellcolor{cyan!20}{0.838(+3.08\%)} & \cellcolor{cyan!20}{0.576(+13.16\%)} \\
\hline
\multirow{3}{*}{\makecell[l]{TCGA-EGFR}} 
& w/o pre-train & 84.54 & 0.737 & 0.540 & 81.53 & 0.613 & 0.816 \\
& \cellcolor{lightgray!50}single-organ 
& \cellcolor{lightgray!50}{85.51(+1.14\%)} & \cellcolor{lightgray!50}{0.743(+0.81\%)} & \cellcolor{lightgray!50}{0.646(+19.63\%)} 
& \cellcolor{lightgray!50}{82.04(+0.63\%)} & \cellcolor{lightgray!50}{0.656(+7.01\%)} & \cellcolor{lightgray!50}{0.821(+0.61\%)}\\
& \cellcolor{cyan!20}multi-organ 
& \cellcolor{cyan!20}{87.44(+3.43\%)} & \cellcolor{cyan!20}{0.771(+4.61\%)} & \cellcolor{cyan!20}{0.670(+24.07\%)} 
& \cellcolor{cyan!20}{83.98(+3.01\%)} & \cellcolor{cyan!20}{0.768(+25.28\%)} & \cellcolor{cyan!20}{0.840(+2.94\%)}\\
\hline
\end{tabular}
\end{table*}

\begin{figure*}[!t]
\centering
    \subfigure[statistic]{
        \includegraphics[width=0.20\linewidth]{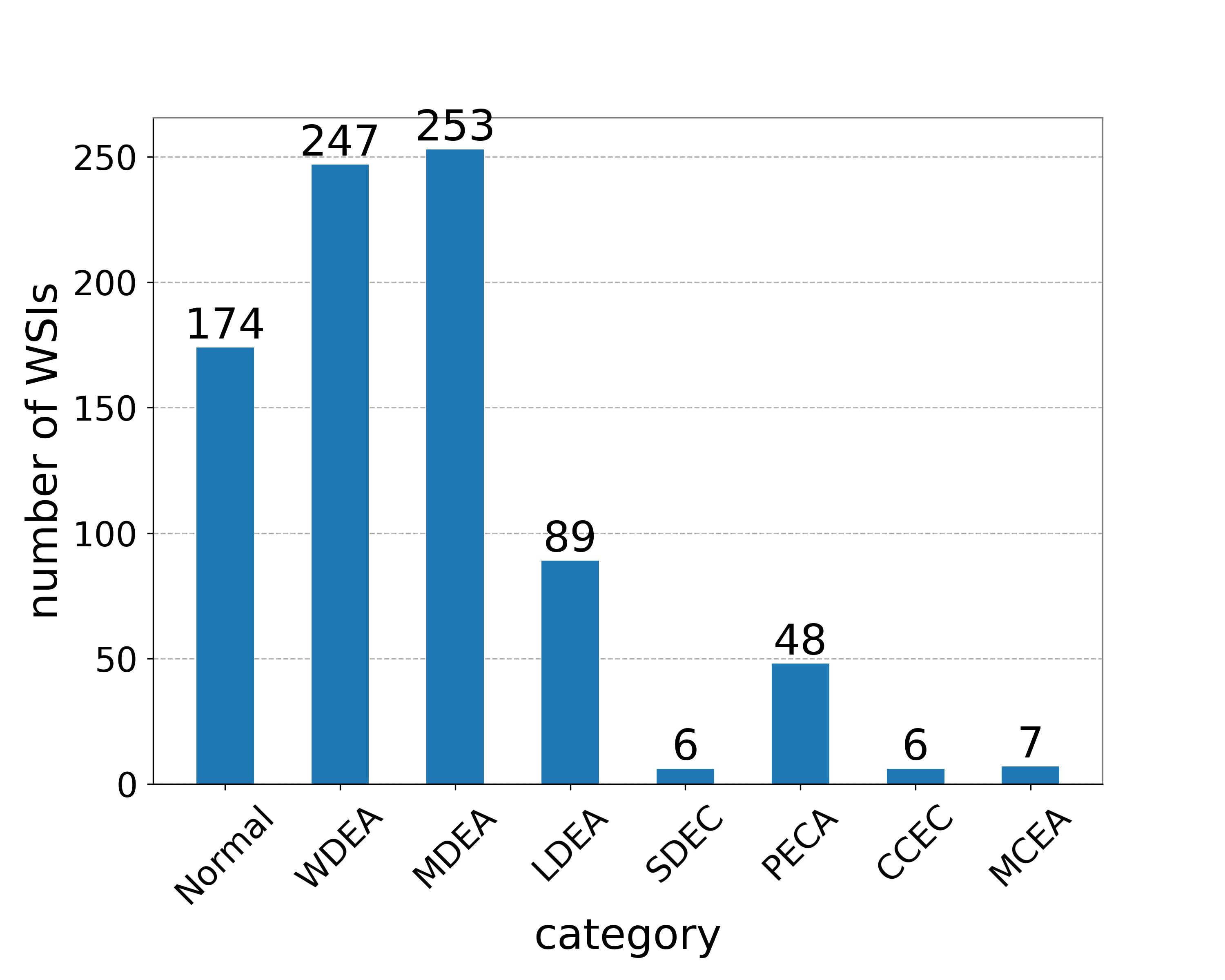}
    }
    \subfigure[w/o pre-train]{
        \includegraphics[width=0.24\linewidth]{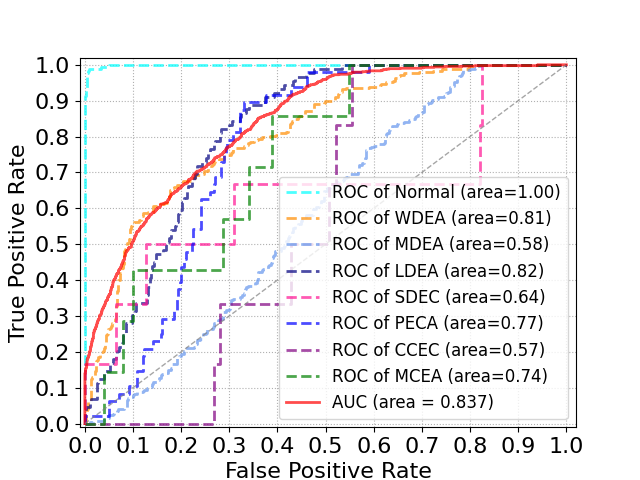}
    } 
    \subfigure[single-organ]{
        \includegraphics[width=0.24\linewidth]{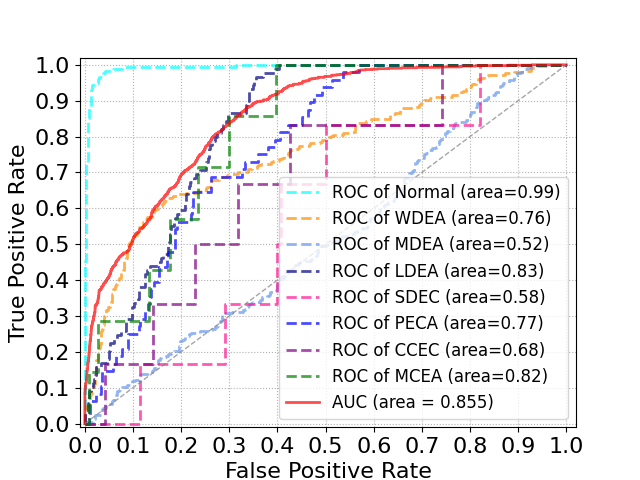}
    } 
    \subfigure[multi-organ]{
        \includegraphics[width=0.24\linewidth]{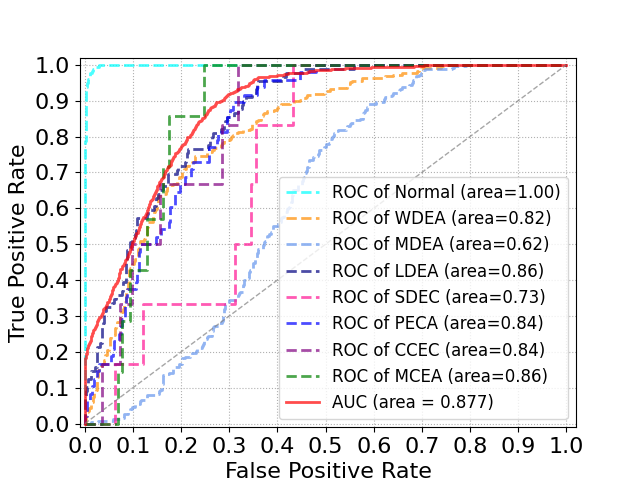}
    }
\caption{Improvement on the long-tailed dataset, where (a) shows the categories distribution of the unbalanced Endometrium-3k dataset, (b), (c), and (d) exhibit the ROC curves of each category without pre-training, with pre-training on the single dataset, and with pre-training on multi-organ datasets using DINO patch features, respectively.}
\label{roc}
\end{figure*}

\subsection{Experimental setting}
During the WSI-level representation pre-training stage, we did not involve any supervised information. The pre-trained encoder will be utilized as the slide representation extractor for various downstream tasks. 
We applied DINO \cite{caron2021emerging} to pre-train and extract all patch features and also utilized the released foundation model PLIP \cite{huang2023visual} as the patch feature extractor on the magnification under 20× lenses. 

We first pre-trained our model on multi-organ datasets and then evaluated the performance on six task-specific datasets with two conditions, where the in-domain condition is that fine-tuning datasets are involved in the pre-training, otherwise is the out-of-domain condition.
Subsequently, we validated the effectiveness of WSI representation learning and conducted comparison experiments with other SOTA methods to showcase the superiority of PAMA. 
In the end, the ablations and parametric experiments demonstrate the significance of the proposed modules and strategy. Accuracy (ACC), the area under the ROC curve (AUC), and the F1 score were used as metrics to evaluate performance.

We implemented all the methods in Python 3.8 with PyTorch 1.7 and Cuda 10.2. Our experiments were conducted on a computer cluster with ten Nvidia Geforce 2080Ti GPUs.

\begin{figure*}[t]
\begin{center}
		\subfigure[AUCs on various downstream tasks.]{
			\includegraphics[width=\linewidth]{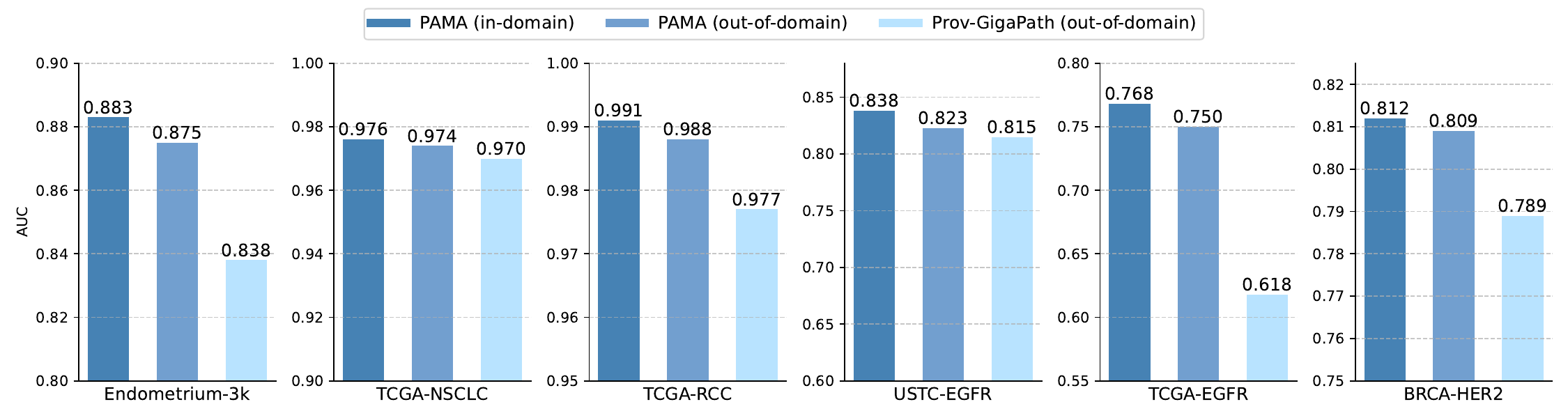}
		}
		\subfigure[ACCs(\%) on various downstream tasks.]{
			\includegraphics[width=\linewidth]{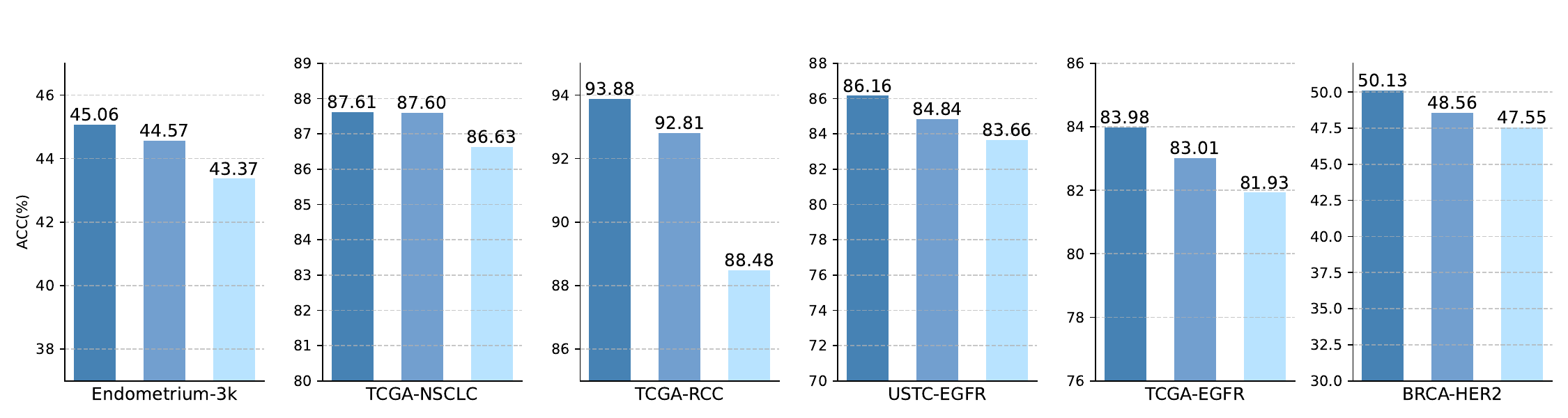}
            }

\caption{Performance of different pre-training conditions of PAMA and Prov-GigaPath on various downstream tasks.}
\label{out_of_domain}
\end{center}
\end{figure*}

\subsection{Effectiveness for in-domain pre-training}
In this experiment, we pre-trained PAMA within the training sets of five datasets, i.e., TCGA-RCC, TCGA-NSCLC, USTC-EGFR, Endometrium-3k, and TCGA-EGFR, which are regarded as in-domain datasets.
Then, we evaluated the performance of the pre-trained model on the test sets of the in-domain datasets to show the effectiveness of PAMA in learning representation from abundant unlabeled histopathology image data.

Table. \ref{table1} shows the results with diffident training strategies using DINO \cite{caron2021emerging} patch features and PLIP \cite{huang2023visual} patch features, where \textit{w/o pre-train} means to directly train the PAMA encoder in a weakly supervised way, \textit{single-organ} refers to pre-training PAMA on the current single dataset and then fine-tuning it with task labels, and \textit{multi-organ} refers to pre-training PAMA on the multi-organ dataset and then fine-tuning it with task-specific labels. 

For every dataset under DINO features, pre-training on the single dataset can increase ACCs by 0.17\% to 22.75\%, increase AUCs by 0.31\% to 2.15\%, and increase F1 score by 0.33\% to 19.63\% for different tasks. 
Multi-organ pre-training further promotes the performance of the model. Specifically, the ACCs/AUCs increase by 29.61\%/4.77\%, 8.49\%/1.85\%, 0.81\%/1.12\%, 5.84\%/2.74\%, and 3.43\%/4.61\% on the Endometrium-3k dataset, TCGA-NSCLC dataset, TCGA-RCC dataset, USTC-EGFR dataset, and TCGA-EGFR dataset, respectively. 
As for the PLIP features, pre-training on the multi-organ datasets can increase ACCs by 1.36\% to 17.25\%, increase AUCs by 0.93\% to 25.28\%, and increase F1 score by 1.62\% to 17.44\% for different tasks.
These results demonstrate the proposed method can effectively promote the WSI encoder in optimizing the use of abundant unlabeled WSI data and enhancing representational abilities. 

Our model gains even more significant improvement on the Endometrium-3k dataset, where the data is extremely unbalanced. Fig. \ref{roc}(a) exhibits that the data of LDEA and PECA are less than half of MEDA data, while SDEA, CCEA, and MCEA are even less than ten WSIs. Datasets with long-tailed categories often lead to model bias problems.
Fig. \ref{roc}(d) shows that multi-organ pre-training increased AUCs by 0.04, 0.09, 0.07, 0.27, and 0.12 for categories of LDEA, SDEC, PECA, CCEC, and MCEA, respectively, when compared with the direct training. It demonstrates that PAMA pre-trained on multiple organ datasets can enhance the model generalization ability to significantly relieve the model bias problem.

Molecular characterizations manifest as more latent features that are not visible in histopathology images, and thus molecular status prediction by WSIs is a more challenging task. Pre-training on the single dataset improves the ACCs/F1 score on the USTC-EGFR dataset and TCGA-EGFR dataset by 3.90\%/6.88\% and 1.14\%/19.63\%. It demonstrates that WSI-level self-supervised learning can obtain more discriminative implicit semantic features. Furthermore, directly fine-tuning the multi-organ pre-trained model on the two datasets contributes to an increase in F1 scores by 15.78\% and 24.07\%. Such a significant improvement indicates that multi-organ pre-training can mine the general semantic information of histopathology images, and thereby can complete various challenging tasks more effectively and efficiently. This demonstrates the ability of our proposed method to be more practical and meaningful in building computer-aided pan-cancer diagnosis systems.

\subsection{Generalization for out-of-domain pre-training}
We additionally collected a large-scale pan-cancer dataset from TCGA as the out-of-domain data to evaluate the generalization of PAMA pre-training. We pre-trained PAMA and a SOTA method, namely Prov-GigaPath \cite{xu2024whole}, on the pan-cancer dataset without any labels, and then fine-tuned the encoder on six downstream tasks completely independent of the pan-cancer dataset. The results are represented as \textit{PAMA (out-of-domain)} and \textit{Prov-GigaPath (out-of-domain)} in Fig.  \ref{out_of_domain}. We conducted the experiment using PLIP \cite{huang2023visual} patch features and the \textit{PAMA (in-domain)} represents the results of \textit{multi-organ} strategy in Table \ref{table1}.

In Fig. \ref{out_of_domain}, the performance of \textit{PAMA (out-of-domain)} decrease by no more than 0.02 in AUCs and no more than 1.6\% in ACCs compared with \textit{PAMA (in-domain)}. Especially on TCGA-NSCLC dataset and TCGA-RCC dataset, the AUCs decreased by less than 0.003, where there is nearly no degradation in performance of pre-training PAMA with the out-of-domain data. The results demonstrate that PAMA exhibits substantial out-of-domain generalization capabilities.

\textit{PAMA (out-of-domain)} are superior to \textit{Prov-GigaPath (out-of-domain)} in AUCs by 0.004 to 0.132 and in ACCs by 0.97\% to 4.33\% for different tasks. It displays a better capacity of our method in characterizing and analyzing unseen data in comparison to Prov-GigaPath.

Furthermore, \textit{PAMA (out-of-domain)} achieves comparable and superior performance compared with pre-training on the single dataset results in Table \ref{table1}. The results effectively demonstrate PAMA's ability to mine information from extensive amounts of unlabeled data, facilitating potential of the framework for more general histopathology image analysis tasks.

\begin{figure}[!t]
\centering{
			\includegraphics[width=\linewidth]{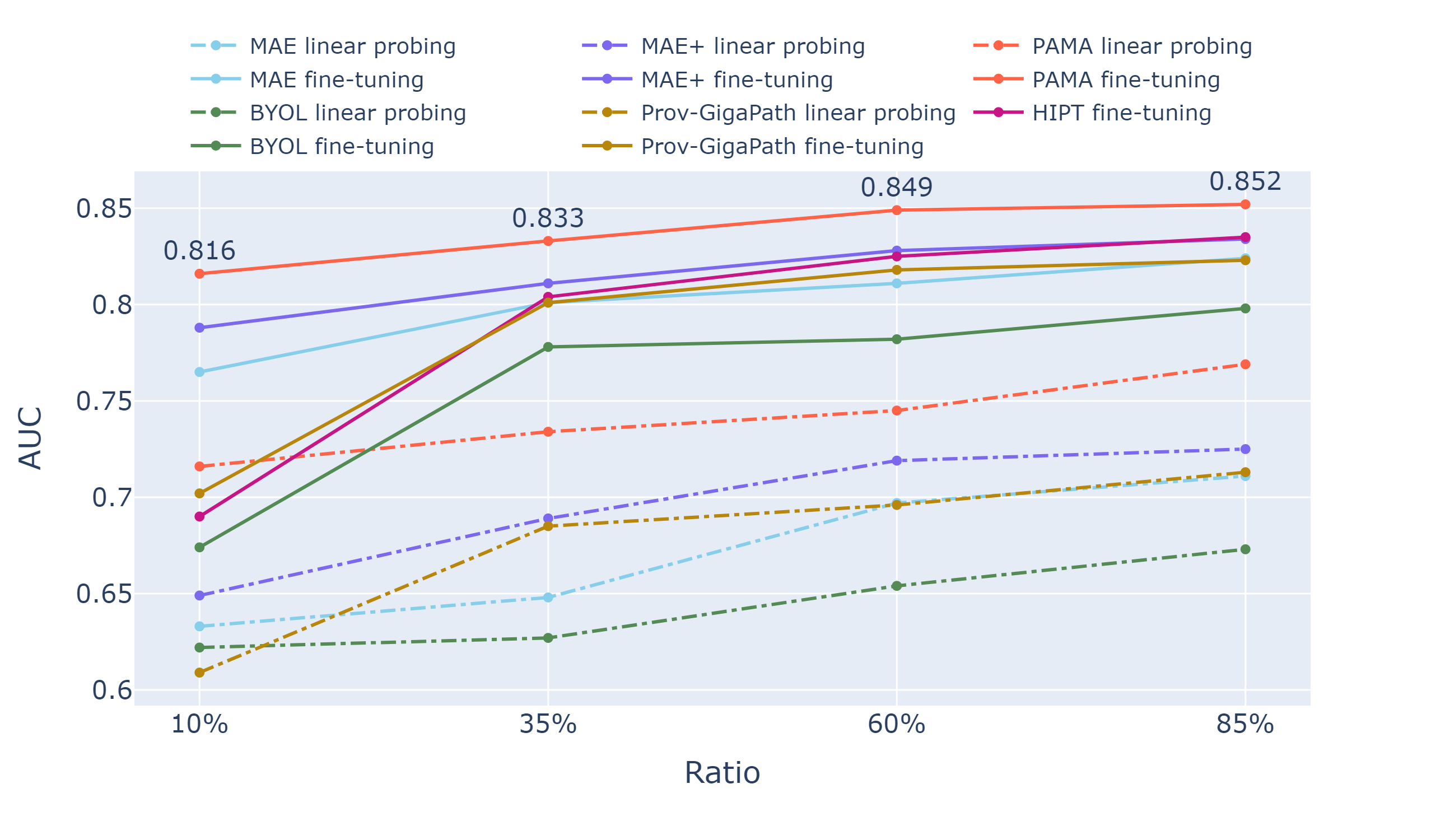}
		}
            
\caption{Semi-supervised experiments on the Endometrium-3k dataset, utilizing 10\%, 35\%, 60\%, and 85\% of labeled data, where fine-tuning results are depicted with solid lines, and linear probing results are denoted with dotted lines.}
\label{semi}
\end{figure}

\subsection{Effectiveness of semi-supervised WSI classification}
Then, we conducted experiments to assess the effectiveness of WSI-level self-supervised learning under conditions with limited WSI labels. The results are presented in Fig. \ref{semi}, which compares the performance obtained with varying ratios of training WSIs with labels.
We re-implemented MAE \cite{he2022masked} for slide-level feature learning as the baseline. Additionally, we applied the proposed distance and polar angle embedding into the self-attention module of MAE, denoted as MAE+ in Fig. \ref{semi}. To ensure the objectivity of the comparisons, we employed the method in the original paper \cite{chen2022scaling} to fine-tune the HIPT. Additionally, we re-implemented BYOL \cite{grill2020bootstrap} as the contrastive learning-based self-supervised slide-level learning for comprehensive comparison with the MIM framework.

It shows that PAMA is consistently superior to MAE, HIPT and Prov-GigaPath \cite{xu2024whole} across all label ratios. Prov-GigaPath utilizes DINO V2 \cite{oquab2023dinov2} to extract patch features and uses LongNet \cite{ding2023longnet} as slide aggregator for pre-training. LongNet was originally designed for extremely long sequences like over 1B+ tokens. However, Prov-GigaPath has not considered any properties of WSI, especially spatial structure information. In sufficient data conditions, it is not even better than HIPT. With the volume of our datasets, the performance of Prov-GigaPath, which is not specifically designed for WSI characteristics, does not differ much from plain MAE, but still has a large margin from PAMA. The above results demonstrate the effectiveness of PAMA in pre-training WSI representations and the MIM frameworks are more efficient than the contrastive leaning framework for slide-level learning.
PAMA obtains optimal stability in AUCs with label ratios reducing from 85\% to 10\%. This is of great practical value as it reduces the reliance on a massive number of labeled WSIs for training a robust WSI analysis model. Meanwhile, we can employ unlabeled WSIs with the assistance of PAMA to enhance the capabilities of the WSI analysis models. HIPT is a two-stage pre-training model that is slightly less effective than the one-stage MAE. This illustrates that the discontinuous gradient back-propagation of a multi-stage pre-training model led to an accumulation of biases. In addition, the MAE+ outperforms MAE. It indicates our proposed distance and polar angle embedding can capture more complete spatial information of WSI than the original position encoding of ViT.

\begin{table*}[t]
\centering
\scriptsize
\caption{Comparison with SOTA WSI analysis methods on the Endometrium-3k dataset.}
\label{table3}
\setlength{\tabcolsep}{8pt}
\begin{tabular}{l|cc|cc|cc|cc|cc}
\hline
\multirow{2}{*}{\textbf{Methods}} & \multicolumn{2}{c|}{10\%} & \multicolumn{2}{c|}{35\%}
                                  & \multicolumn{2}{c|}{60\%} & \multicolumn{2}{c|}{85\%} & \multicolumn{2}{c}{100\%}\\
                                  & AUC & ACC (\%) & AUC & ACC (\%) & AUC & ACC (\%) & AUC & ACC (\%) & AUC & ACC (\%)\\ 
\hline
\hline
DSMIL \cite{li2021dual} & 0.649 & 25.31 & 0.761 & 38.21 & 0.769 & 38.51 & 0.772 & 38.94 & 0.786 & 39.32 \\
TransMIL \cite{shao2021transmil} & 0.661 & 26.74 & 0.783 & 38.43 & 0.788 & 38.91 & 0.795 & 39.51 & 0.798 & 40.01 \\
SETMIL \cite{zhao2022setmil}& 0.685 & 27.56 & 0.795 & 38.71 & 0.810 & 38.89 & 0.829 & 40.08 & 0.831 & 40.84 \\
KAT \cite{zheng2023kernel} & 0.688 & 27.61 & 0.799 & 38.89 & 0.817 & 39.02 & 0.831 & 40.72 & 0.835 & 41.93 \\
\hline
BYOL \cite{grill2020bootstrap} & 0.674 & 27.95 & 0.778 & 38.25 & 0.782 & 38.34 & 0.798 & 38.79 & 0.812 & 40.04 \\
HIPT \cite{chen2022scaling} & 0.690 & 28.68 & 0.804 & 38.69 & 0.825 & 39.19 & 0.835 & 41.09 & 0.842 & 40.63 \\
MAE \cite{he2022masked} & 0.765 & 37.69 & 0.801 & 38.87 & 0.811 & 39.54 & 0.824 & 39.96 & 0.832 & 41.95 \\
Prov-GigaPath \cite{xu2024whole} & 0.702 & 33.25 & 0.801 & 38.67 & 0.818 & 39.75 & 0.823 & 41.56 & 0.839 & 42.28 \\
\hline
PAMA & \textbf{0.816} & \textbf{43.18} & \textbf{0.833} & \textbf{44.94} & \textbf{0.849} & \textbf{45.72} & \textbf{0.852} & \textbf{46.96} & \textbf{0.855} & \textbf{47.47} \\
\hline
\end{tabular}
\caption{Comparison with SOTA WSI analysis methods on the TCGA-NSCLC dataset.}
\label{table4}
\begin{tabular}{l|cc|cc|cc|cc|cc}
\hline
\multirow{2}{*}{\textbf{Methods}}  & \multicolumn{2}{c|}{10\%} & \multicolumn{2}{c|}{35\%}
                                  & \multicolumn{2}{c|}{60\%} & \multicolumn{2}{c|}{85\%} & \multicolumn{2}{c}{100\%}\\
                                  & AUC & ACC (\%) & AUC & ACC (\%) & AUC & ACC (\%) & AUC & ACC (\%) & AUC & ACC (\%)\\
\hline
\hline
DSMIL \cite{li2021dual} & 0.833 & 67.50 & 0.911 & 75.00 & 0.921 & 77.71 & 0.931 & 78.04 & 0.938 & 80.11  \\
TransMIL \cite{shao2021transmil} & 0.867 & 69.01 & 0.932 & 79.62 & 0.941 & 80.28 & 0.949 & 81.49 & 0.959 & 84.35  \\
SETMIL \cite{zhao2022setmil} & 0.891 & 72.71 & 0.937 & 80.21 & 0.945 & 81.05 & 0.953 & 82.47 & 0.962 & 84.95  \\
KAT \cite{zheng2023kernel} & 0.915 & 76.01 & 0.951 & 83.37 & 0.954 & 83.57 & 0.957 & 83.68 & 0.965 & 85.81  \\
\hline
BYOL \cite{grill2020bootstrap} & 0.876 & 71.95 & 0.912 & 76.52 & 0.943 & 79.78 & 0.955 & 83.37 & 0.964 & 84.45 \\
HIPT \cite{chen2022scaling} & 0.948 & 80.90 & 0.967 & 84.23 & 0.970 & 85.36 & 0.975 & 86.57 & 0.977 & 87.83 \\
MAE \cite{he2022masked} & 0.951 & 82.28 & 0.965 & 83.90 & 0.966 & 84.64 & 0.968 & 85.54 & 0.970 & 87.50  \\
Prov-GigaPath \cite{xu2024whole} & 0.927 & 78.04 & 0.962 & 85.65 & 0.964 & 86.19 & 0.967 & 86.63 & 0.974 & 89.89 \\
\hline
PAMA & \textbf{0.978} & \textbf{89.02} & \textbf{0.984} & \textbf{91.74} & \textbf{0.985} & \textbf{91.87} & \textbf{0.987} & \textbf{92.39} & \textbf{0.988} & \textbf{92.72} \\
\hline
\end{tabular}
\end{table*}

\subsection{Comparison with other weakly supervised methods}
We compared PAMA with four self-supervised frameworks, BYOL, MAE, HIPT, and Prov-GigaPath, and four SOTA weakly supervised methods, including DSMIL \cite{li2021dual}, TransMIL \cite{shao2021transmil}, SETMIL \cite{zhao2022setmil}, and KAT \cite{zheng2023kernel} on the Endometrium-3k and TCGA-NSCLC datasets for slide-level classification. The results are shown in Table \ref{table3} and \ref{table4}.

Overall, our proposed PAMA is superior to the second-best method with increased AUCs/ACCs(\%) of 0.051/5.49, 0.029/6.05, 0.024/5.97, 0.017/5.40, and 0.013/5.19 on the Endometrium-3k dataset with 10\%, 35\%, 65\%, 85\% and 100\% labeled data, and increased AUCs/ACCs(\%) of 0.027/6.74, 0.017/6.09, 0.015/5.68, 0.012/5.76, and 0.011/2.83 on the TCGA-NSCLC dataset with 10\%, 35\%, 65\%, 85\% and 100\% labeled training data, respectively.

DSMIL \cite{li2021dual} introduced a dual-stream architecture with trainable distance measurement for instances and applied a pyramidal fusion framework for multiscale WSI features, which, however, did not consider the spatial structure of tissue. The absolute structural encoding reduces the performance of DSMIL from other methods. TransMIL and SETMIL leveraged CNN blocks to aggregate local information and then built Transformer structures for long-range global feature aggregation. KAT considered the spatial adjacency of patches and manually defined the fixed hierarchical masks based on local kernels to maintain relative distance information. None of the three methods embedded relative orientation information into slide representations, which causes a significant performance gap compared with PAMA. We re-implemented BYOL \cite{grill2020bootstrap} with ViT \cite{dosovitskiy2020image} backbone for slide-level feature learning based contrastive learning. Contrastive self-supervised learning frameworks like BYOL rely on extensive and effective augmented views to mine the discriminative representations. However, there are no efficient published WSI-level view augmentation methods currently and we applied random sample patches to construct different views of the WSI. WSI views based on random patch sampling are struggling to efficiently capture semantic information. It results in even worse performance than some weak-supervised methods.
Self-supervised learning methods based on MIM, namely, MAE and Prov-GigaPath, surpass these SOTA weakly supervised methods. It reconfirms the effectiveness of WSI-level representation pre-training.

Additionally, PAMA fine-tuned on 35\% labeled data on the two datasets can achieve comparable results with other methods trained on 100\% labeled data. It demonstrates that PAMA is capable of utilizing limited data more effectively, decreasing the reliance on large amounts of labeled data for training high-capacity models.

\subsection{Ablation studies}

\begin{table}[!t]
\centering
\caption{Ablation studies on the Endometrium-3k dataset.}
\label{table2}
\setlength{\tabcolsep}{5pt}
\begin{tabular}{c|cccc|lll}
\hline
\textbf{NO.}& \textbf{Dis}& \textbf{Polar}& \textbf{KRO} & \textbf{AD} & AUC & ACC (\%)  \\
\hline
1 & $\checkmark$ & $\checkmark$ & $\checkmark$ & $\checkmark$ & 0.855 & 47.47 \\

2 & $\checkmark$ & $\checkmark$ & $\checkmark$ &  &0.851 ($\downarrow$0.004) & 43.64 ($\downarrow$3.83) \\

3 & $\checkmark$ & $\checkmark$ &  & $\checkmark$ & 0.826 ($\downarrow$0.029) & 40.60 ($\downarrow$6.87)\\

4 & $\checkmark$ &  &  & $\checkmark$ & 0.839 ($\downarrow$0.016) & 41.08 ($\downarrow$6.39) \\

5 &  & $\checkmark$ & $\checkmark$ & $\checkmark$ & 0.833 ($\downarrow$0.022) & 40.72 ($\downarrow$6.75) \\

6 &  &  &  & $\checkmark$ & 0.821 ($\downarrow$0.034) & 39.51 ($\downarrow$7.96)\\
\hline
\end{tabular}
\end{table}
We conducted ablation studies on the Endometrium-3k dataset to verify the significance of our relative spatial embedding and strategy shown in Table. \ref{table2}. When the polar angel embedding of anchors was removed, we observed the AUC and ACC(\%) dropped by 0.016 and 6.39. It is notable that if we applied the polar embedding without the KRO module, the AUC and ACC(\%) dropped by 0.029 and 6.87, which means that indexing angles with a fixed polar axis will lead to ambiguous semantic information in WSIs. KRO strategy can dynamically update every anchor’s main polar axis to disambiguate structure information in slides. The relative distance embedding can maintain scale-varying WSIs in a semantic consistency space. The AUC and ACC(\%) decreased by 0.022 and 6.75, respectively, when the distance embedding was discarded. When we constructed the slide representation neither with the distance nor polar angle embeddings, the performance had a significant drop of 0.034/7.96 in the AUC/ACC(\%). These results prove that the proposed modules can effectively and efficiently acquire spatial information to maintain semantic integrity and consistency in WSIs. Furthermore, the AUC and ACC(\%) decreased by 0.004 and 3.83, respectively, when the AD was discarded. This indicates that anchor dropout contributes to better generalization performance.

\begin{figure*}[t]
\begin{center}
		\subfigure[masking ratio $r$]{
			\includegraphics[width=0.2\linewidth]{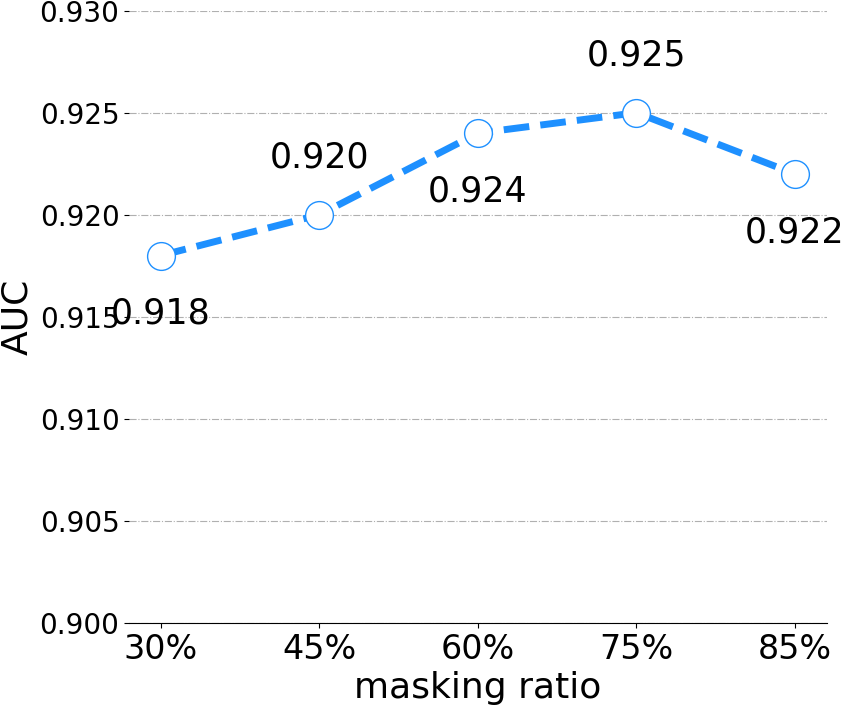}
		}
		\subfigure[drop out probability $p$]{
			\includegraphics[width=0.2\linewidth]{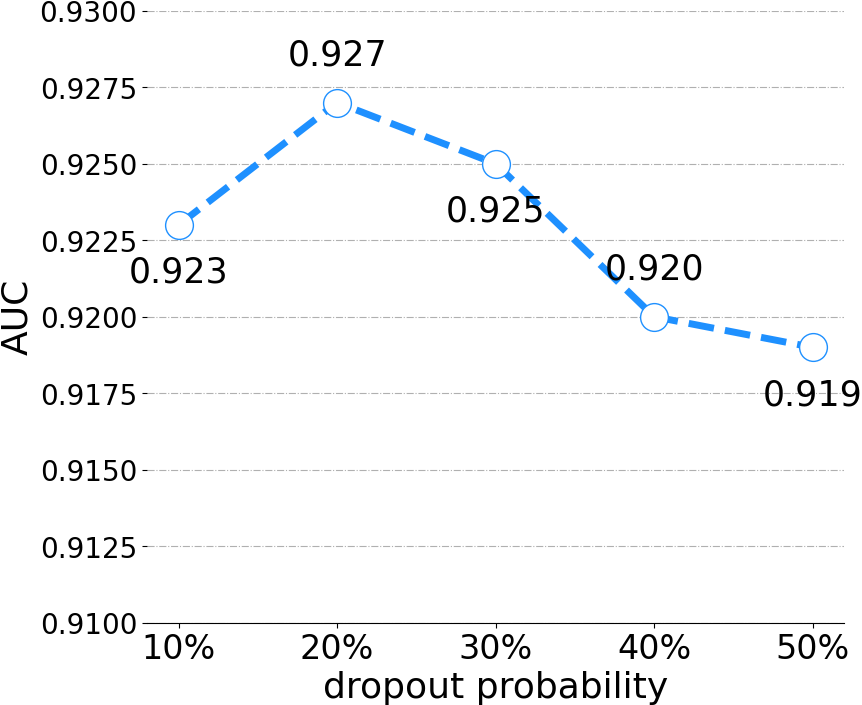}
		} 
            \subfigure[patches per anchor $c$]{
			\includegraphics[width=0.2\linewidth]{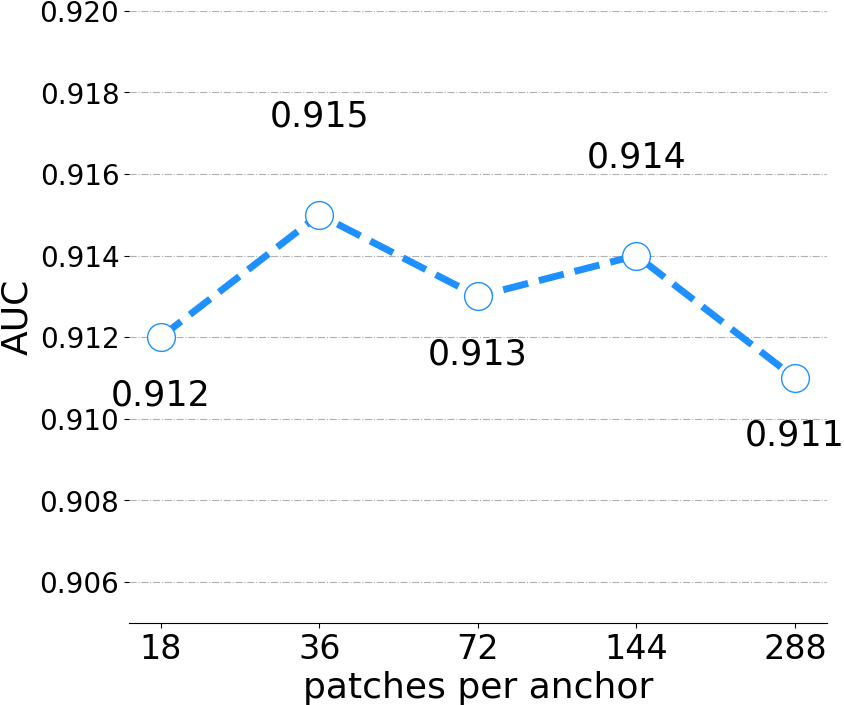}
		}
		\subfigure[polar bins $N$]{
			\includegraphics[width=0.2\linewidth]{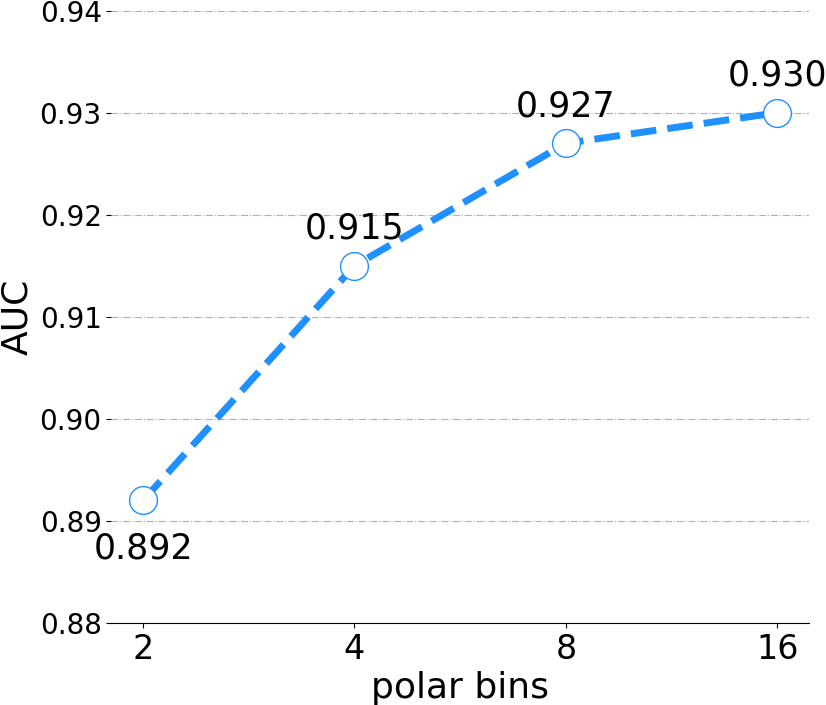}
		}
\caption{Performance of PAMA with different hyper-parameter settings on the Endometrium-3k validation set.}
\label{hyperparameter}
\end{center}
\end{figure*}

\subsection{Hyperparameter analysis}
To verify the design of the PAMA framework, we performed a series of parametric experiments on the Endometrium-3k dataset, where only the current parameter was tuned in each set of experiments and the remaining parameters were fixed. The results are shown in Fig. \ref{hyperparameter}.
\begin{enumerate}
    \item \textit{Masking ratio}: $r$ is the ratio of masked patches to remove before we feed the remaining tokens into the PAMA encoder during pre-training. We find that masking nearly 75\% tokens to reconstruct the slide representation can help the model obtain a promising performance in Fig. \ref{hyperparameter}(a). Reducing the masking ratio limits the model's reconstruction space, whereas an excessively high ratio sacrifices fundamental contextual information.
    \item \textit{Dropout probability}: $p$ is the probability of randomly discarding anchors for data augmentation. Different reserved anchors can lead to the diverse structural representations of the WSI. Fig. \ref{hyperparameter}(b) indicates that PAMA with dropout $20\%$ anchors achieves the best performance. The model performs stable when the probability is higher than $30\%$, which demonstrates that discarding a wide range of anchors will cause the basic information of the WSI to be missed.
    \item \textit{Patches per anchor}: $c$ denotes the number of patches per anchor clustering cluster. Increasing the value of $c$ will enable the anchor to capture a wider range of contextual information, whereas reducing its value will result in the generation of more anchors which means a higher computational amount. Based on Fig. \ref{hyperparameter}(c), we set $c=144$ for balancing performance and resource consumption.
    \item \textit{Polar bins}: $N$ is the number of orientation bins, $e.g.$, $N=8$ means each bin holds a $\frac{2\pi}{N}=\frac{\pi}{4}$ angle range. As $N$ increases, the anchor can provide more precise structural information due to the detailed division of orientation intervals. However, this enhancement comes at the cost of increased computational consumption. Based on Fig. \ref{hyperparameter}(d), we set $N=8$ for balancing performance and resource consumption.
\end{enumerate}

\begin{figure*}[!t]
\centering
\includegraphics[width=0.87\linewidth]{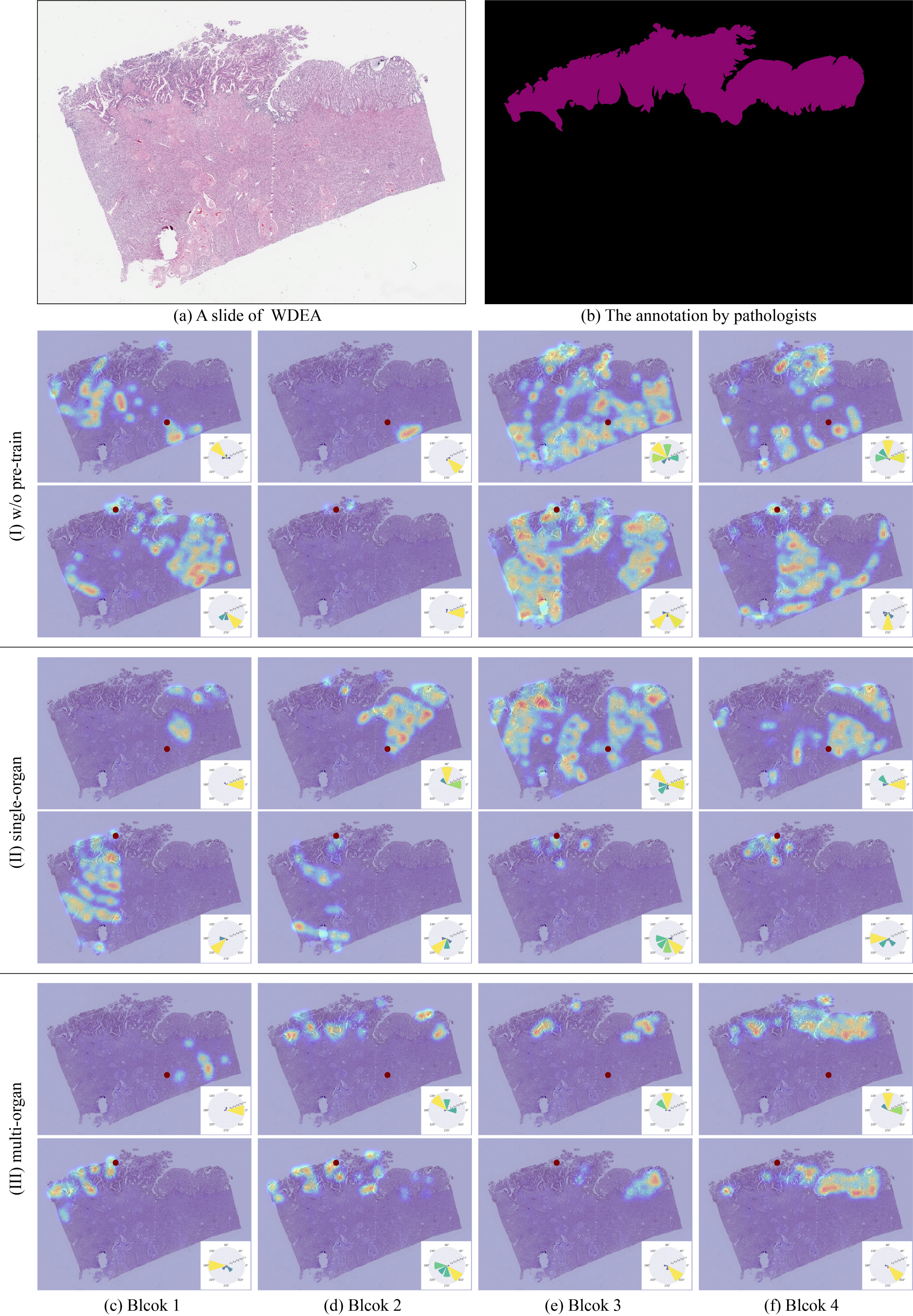}
\caption{The visualization of the anchor attention in the PACA module without pre-train and during fine-tuning after pre-training with single-organ and multi-organ datasets, where (a) showcases a well-differentiated endometrioid adenocarcinoma slide, (b) is the annotation by pathologists, and (c)-(f) show the attention heatmaps based on anchors in each PACA block in which the selected anchor positions are indicated by red dots and the polar attention distribution is shown in the bottom right corner of each diagram.}
\label{visualization2}
\end{figure*}

\begin{figure*}[!t]
\centering
\includegraphics[width=0.87\linewidth]{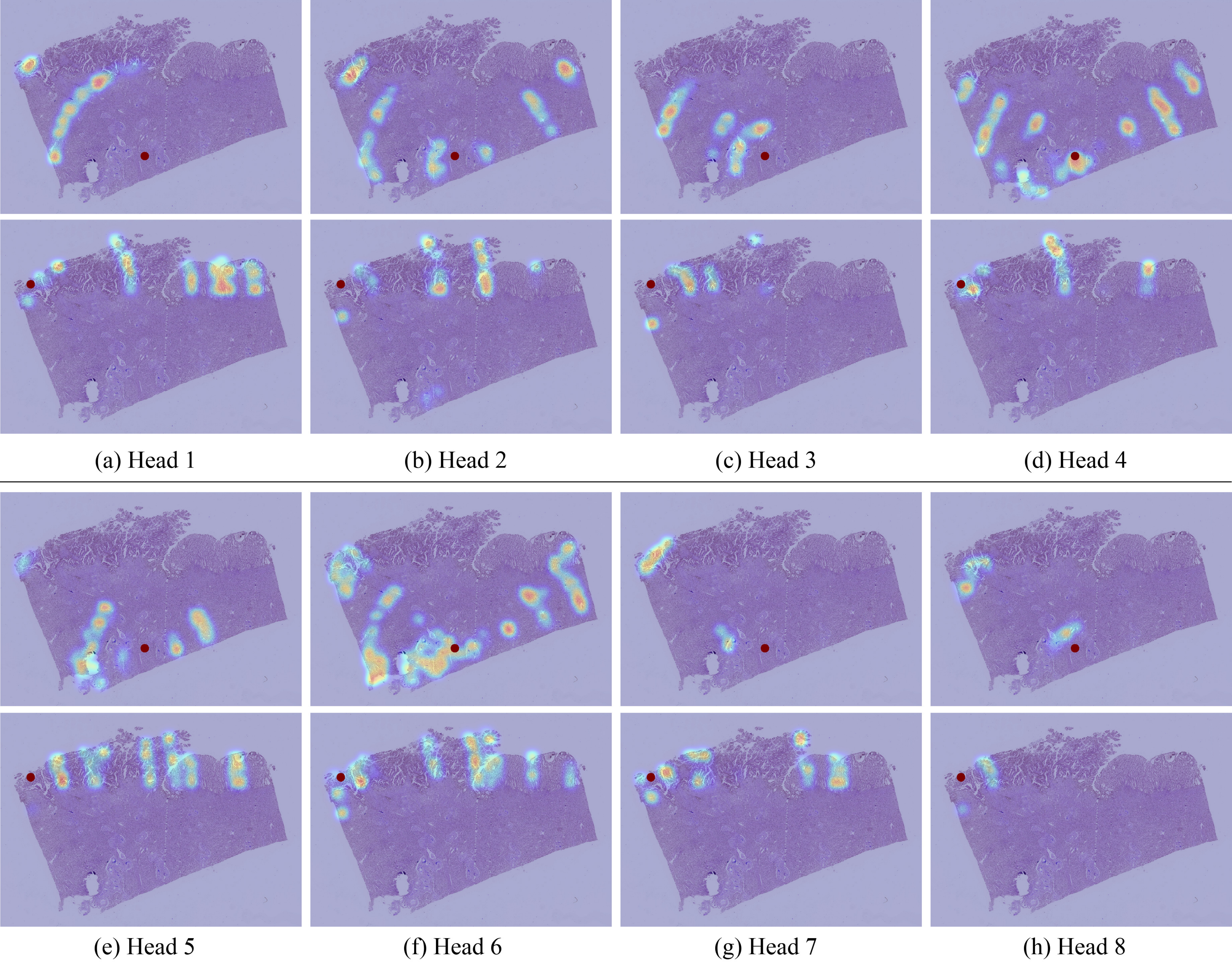}
\caption{The visualization of multi-head attention based on anchors in the PACA during pre-training, where the anchor in the top row is located within the non-cancerous tissue region while the anchor in the bottom row is located in the cancerous tissue region.}
\label{visualization1}
\end{figure*}

\section{Visualization}
We further assessed the interpretability of our proposed framework with visualization. We present a well-endometrioid adenocarcinoma slide and the annotation by pathologists as shown in Fig. \ref{visualization2}(a-b). Fig. \ref{visualization2}(c-f) show the heatmap and polar attention distribution based on anchors in each PACA block without pre-train and during fine-tuning after pre-training with single-organ and multi-organ datasets. In the early stage ($e.g.$, block 1) during fine-tuning after multi-organ pre-training as shown in Fig. \ref{visualization2}(III), anchors initially focus on identical pathological tissues as the observation regions. Through supervised WSI label fine-tuning, the anchor's attention consolidates on high-risk cancerous tissues and attains stability in which the KRO strategy takes a crucial role in adaptively updating the polar axis that is illustrated by the yellow sector in the radar chart. In the process without pre-train as shown in Fig. \ref{visualization2}(I), the regions of interest of both anchors in the normal and cancerous tissues are diffuse. After pre-training with the single-organ dataset as shown in Fig. \ref{visualization2}(II), the anchor in the positive area can gradually converge to the cancerous tissues. With the contribution of multi-organ pre-training as shown in Fig. \ref{visualization2}(III), anchors' areas of interest are more comprehensive and precise.

Fig. \ref{visualization1} exhibits the multi-head attention heatmap based on anchors during multi-organ pre-training. 
We observe that an anchor located in the negative region is assigned a higher attention score to negative tissue, whereas a positive anchor is given greater attention to cancerous tissue, which means the anchors focus on tissues that share similarities with their features. This behavior enables PAMA to comprehensively describe patterns in histopathology images. From the perspective of the heads, some heads focus on more sparse areas ($e.g.$, head 7 and head 8), while others concentrate on more dense areas ($e.g.$, head 5 and head 6). It is observed that the distance and polar angle range of each head's attention varies and complements each other. This demonstrates that our proposed anchor-based cross-attention module can obtain diverse semantic information without introducing supervision.

\section{Discussion}
Most of the current large-scale pathology foundation models focused on patch-level representation learning \cite{huang2023visual, lu2024visual, ikezogwo2024quilt, chen2024towards, vorontsov2024foundation}. A few works focused on slide-level foundation models, such as HIPT \cite{chen2022scaling} and Prov-GigaPath \cite{xu2024whole}, but they disregarded the properties of WSIs, especially the complex spatial semantic information. We introduced the spatial semantic completeness of WSI into the pre-training process, enhancing the slide representations of PAMA to become more semantically complete and generalized.

Data-driven pre-training strategy following foundation models \cite{huang2023visual, lu2024visual, chen2024towards, xu2024whole, vorontsov2024foundation} facilitated PAMA for pan-cancer analysis. In this paper, we focused on model design and pan-cancer dataset construction. The proposed position-aware cross-attention model with a dynamical reorientation strategy captures the intrinsic semantic representation of WSIs across various cancer types rather than focusing on any specific tumor or organ. Moreover, the framework was pre-trained to obtain the morphological consistency across multiple cancers based on the broad data including over 13.6k WSIs of 22 cancer types covering 11 organs from multiple medical centers. In future work, it will be necessary to further investigate the spatial properties of pan-cancer and to employ explicit designs to mine its semantic information, such as constructing loss function.

We evaluated the generalization of PAMA across various downstream tasks, including tumor sub-typing, gene mutation prediction, and biomarker status grading. Additionally, out-of-domain datasets were constructed to further demonstrate that the pre-trained model can generalize to datasets not included in the pre-training process. Technically, our pre-training process is task-agnostic, allowing the model to be fine-tuned for specific tasks on any downstream task based on pathology WSIs, which is similar to the released foundation pre-training models \cite{xu2024whole, chen2024towards, lu2024visual, vorontsov2024foundation}. Furthermore, we will explore more general downstream tasks for histopathology image analysis based on the PAMA pre-trained model to facilitate its clinical adoption value.

\section{Conclusion}
In the paper, we focused on self-supervised WSI-level representation learning and proposed the position-aware masked autoencoder (PAMA) for WSI pre-training. The proposed anchor-based position-aware cross-attention (PACA) module leverages the bidirectional communication between the local and global information to capture WSI semantic features. We also introduced a kernel reorientation (KRO) strategy to dynamically update the main orientation of anchors to eliminate ambiguity for WSI representation learning. Additionally, we collected seven large-scale datasets from multiple organs and evaluated the effectiveness and generalization of PAMA for pan-cancer analysis. The comprehensive experimental results have demonstrated that the proposed method is superior to the state-of-the-art methods and efficiently facilitates the analysis of pan-cancer. The current work has two limitations that can be improved: (1) the collected multi-organ datasets do not yet contain comprehensive cancer types and need to be further expanded for pan-cancer analysis, and (2) the PAMA structure currently relies only on pathology image data, and we need to further introduce multimodal data, such as genomics, to participate in the pre-training process to facilitate cancer diagnosis. In the future work, we will focus on these challenges to enhance our work.

\textbf{Acknowledgments.} This work was partly supported by Beijing Natural Science Foundation (Grant No. 7242270), partly supported by the National Natural Science Foundation of China (Grant No. 62171007, 61901018, and 61906058), partly supported by the Fundamental Research Fund for the Central Universities of China (grant No. YWF-23-Q-1075), partly supported by the Anhui Provincial Natural Science Foundation (Grant No. 2408085MF162), partly supported by Emergency Key Program of Guangzhou Laboratory (Grant No. EKPG21-32), partly supported by Joint Fund for Medical Artificial Intelligence (Grant No. MAI2023C014), and partly supported by National Key Research and Development Program of China (Grant No. 2021YFF1201004). 


% \bibliographystyle{IEEEtran}
% % argument is your BibTeX string definitions and bibliography database(s)
% \bibliography{bib/IEEE}

\end{document}